\documentclass[sigconf]{acmart}
\AtBeginDocument{%
  }
\copyrightyear{2025}
\acmYear{2025}
\setcopyright{acmlicensed}
\acmConference[MM '25] {Proceedings of the 33rd ACM International Conference on Multimedia}{October 27--31, 2025}{Dublin, Ireland.}
\acmBooktitle{Proceedings of the 33rd ACM International Conference on Multimedia (MM '25), October 27--31, 2025, Dublin, Ireland}
\acmDOI{10.1145/3746027.3755356}
\acmISBN{979-8-4007-2035-2/2025/10}

\usepackage{enumitem}
\usepackage[linesnumbered,ruled,vlined]{algorithm2e}
\usepackage{multirow}
\usepackage{amsthm}
\usepackage{float}
\usepackage{arydshln}
\usepackage{tablefootnote}
\usepackage{stfloats} 

\begin{document}

\title{Enhancing Small-Scale Dataset Expansion with Triplet-Connection-based Sample Re-Weighting}

\author{Ting Xiang}
\authornote{Both authors contributed equally to this research.}
\email{txiang@hnu.edu.cn}
\orcid{0009-0008-1133-490X}
\affiliation{
  \institution{Hunan University}
  \city{Changsha}
  \state{Hunan}
  \country{China}
}

\author{Changjian Chen}
\authornotemark[1]
\email{changjianchen@hnu.edu.cn}
\orcid{0000-0003-2715-8839}
\affiliation{
  \institution{Hunan University}
  \city{Changsha}
  \state{Hunan}
  \country{China}
}

\author{Zhuo Tang}
\authornote{Corresponding author.}
\email{ztang@hnu.edu.cn}
\orcid{0000-0001-9081-8153}
\affiliation{
  \institution{Hunan University}
  \city{Changsha}
  \state{Hunan}
  \country{China}
}

\author{Qifeng Zhang}
\email{zqfhnucsee@hnu.edu.cn}
\orcid{0009-0008-0906-4509}
\affiliation{
  \institution{Hunan University}
  \city{Changsha}
  \state{Hunan}
  \country{China}
}

\author{Fei Lyu}
\email{feilv@hnu.edu.cn}
\orcid{0009-0002-6609-6247}
\affiliation{
  \institution{Hunan University}
  \city{Changsha}
  \state{Hunan}
  \country{China}
}

\author{Li Yang}
\email{yanglixt@hnu.edu.cn}
\orcid{0000-0002-8929-7554}
\affiliation{
  \institution{Hunan University}
  \city{Changsha}
  \state{Hunan}
  \country{China}
}

\author{Jiapeng Zhang}
\email{zhangjp@hnu.edu.cn}
\orcid{0000-0002-0364-3568}
\affiliation{
  \institution{Hunan University}
  \city{Changsha}
  \state{Hunan}
  \country{China}
}

\author{Kenli Li}
\email{lkl@hnu.edu.cn}
\orcid{0000-0002-2635-7716}
\affiliation{
  \institution{Hunan University}
  \city{Changsha}
  \state{Hunan}
  \country{China}
}

\renewcommand{\shortauthors}{Ting Xiang et al.}
\def \sys {TriReWeight}
\def \etal {{et al.\thinspace}}
\newcommand{\eg}{\emph{e.g.}}
\newcommand{\ie}{\emph{i.e.}}

\newcommand{\myparagraph}[1]{\noindent\textbf{#1}}
\newcommand{\xiangting}[1]{\textcolor{black}{#1}}
\newcommand{\changjian}[1]{\textcolor{black}{#1}}
\newcommand{\doubleunderline}[1]{%
  \underline{\underline{#1}}%
}

\begin{abstract}
The performance of computer vision models in certain real-world applications, such as medical diagnosis, is often limited by the scarcity of available images.
Expanding datasets using pre-trained generative models is an effective solution.
However, due to the uncontrollable generation process and the ambiguity of natural language,
noisy images may be generated.
Re-weighting is an effective way to address this issue by assigning low weights to such noisy images.
We first theoretically analyze three types of supervision for the generated images.
Based on the theoretical analysis, we develop {\sys}, a triplet-connection-based sample re-weighting method to enhance generative data augmentation.
Theoretically, {\sys} can be integrated with any generative data augmentation methods and never downgrade their performance.
Moreover, its generalization approaches the optimal in the order $O(\sqrt{d\ln (n)/n})$.
Our experiments validate the correctness of the theoretical analysis and demonstrate that our method outperforms the existing SOTA methods by $7.9\%$ on average over six natural image datasets and by $3.4\%$ on average over three medical datasets.
We also experimentally validate that our method can enhance the performance of different generative data augmentation methods.
\end{abstract}




\maketitle

\section{Introduction}
\label{sec:intro}
The success of computer vision relies on large numbers of high-quality training images~\cite{deng2009imagenet}.
However, manually collecting and labeling large-scale image datasets is costly and time-consuming~\cite{qi2020small}. 
Moreover, due to data scarcity and privacy protections, acquiring large-scale datasets is even impossible in specific domains, such as medical scenarios~\cite{wang2024autoencoder}. 
Recently, with the advancement of generative models (\eg, Stable Diffusion~\cite{rombach2022high}), generative data augmentation~\cite{zhang2024expanding, azizi2023synthetic, chen2025human, yuan2024diffmat}—also referred to as dataset expansion—has emerged as an effective solution to this challenge, such as GIF~\cite{zhang2024expanding} and DiffuseMix~\cite{islam2024diffusemix}.
Given an original image and its label, these methods perturb the latent space of the generative model or change the text prompts to generate several similar but distinct images.
Then, each generated image is assigned the label of the associated original image.
These data, along with the original ones, are combined to train models for downstream tasks, such as classification, enhancing overall performance.

\begin{figure}
    \centering
    \includegraphics[width=\linewidth]{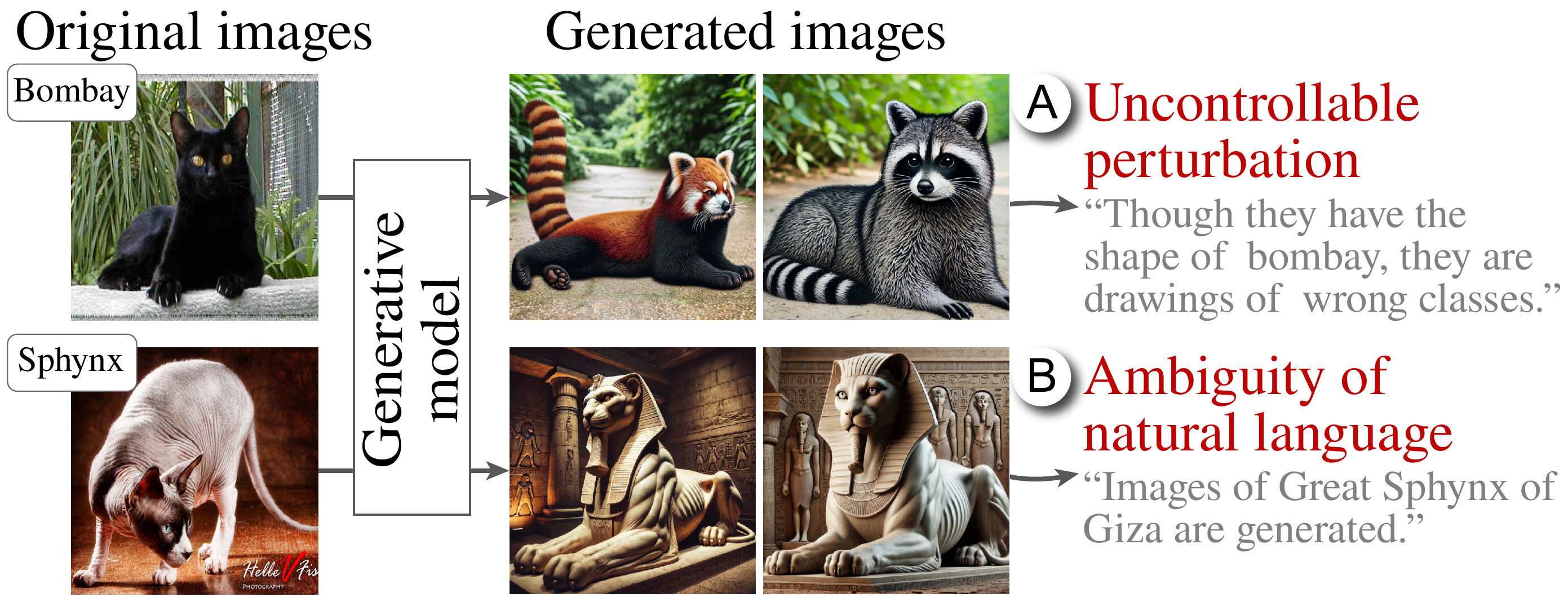}
    \caption{Two exemplar original images and their associated generated images.}
    \Description{}
    \label{fig:samples}
    \vspace{-0.5em}
\end{figure}

Despite the effectiveness of these methods, the model performance is still limited due to the generation of noisy images that are not of interest, arising from two primary factors.
Firstly, since the perturbation in the latent space is uncontrollable, the generation of undesired images is inevitable.
For example, given an image of a Bombay cat as the original image, images of raccoons are generated because their body shapes are similar to the cat (Fig.~\ref{fig:samples}A).
Secondly, due to the ambiguity of natural language,
the text prompt is sometimes interpreted incorrectly.
For example, when generating the images of Sphynx cats, ``Sphynx'' is incorrectly interpreted as ``\textit{Great Sphynx of Giza}'', resulting in wrongly generated images (Fig.~\ref{fig:samples}B).


\begin{figure}
    \centering
    \includegraphics[width=\linewidth]{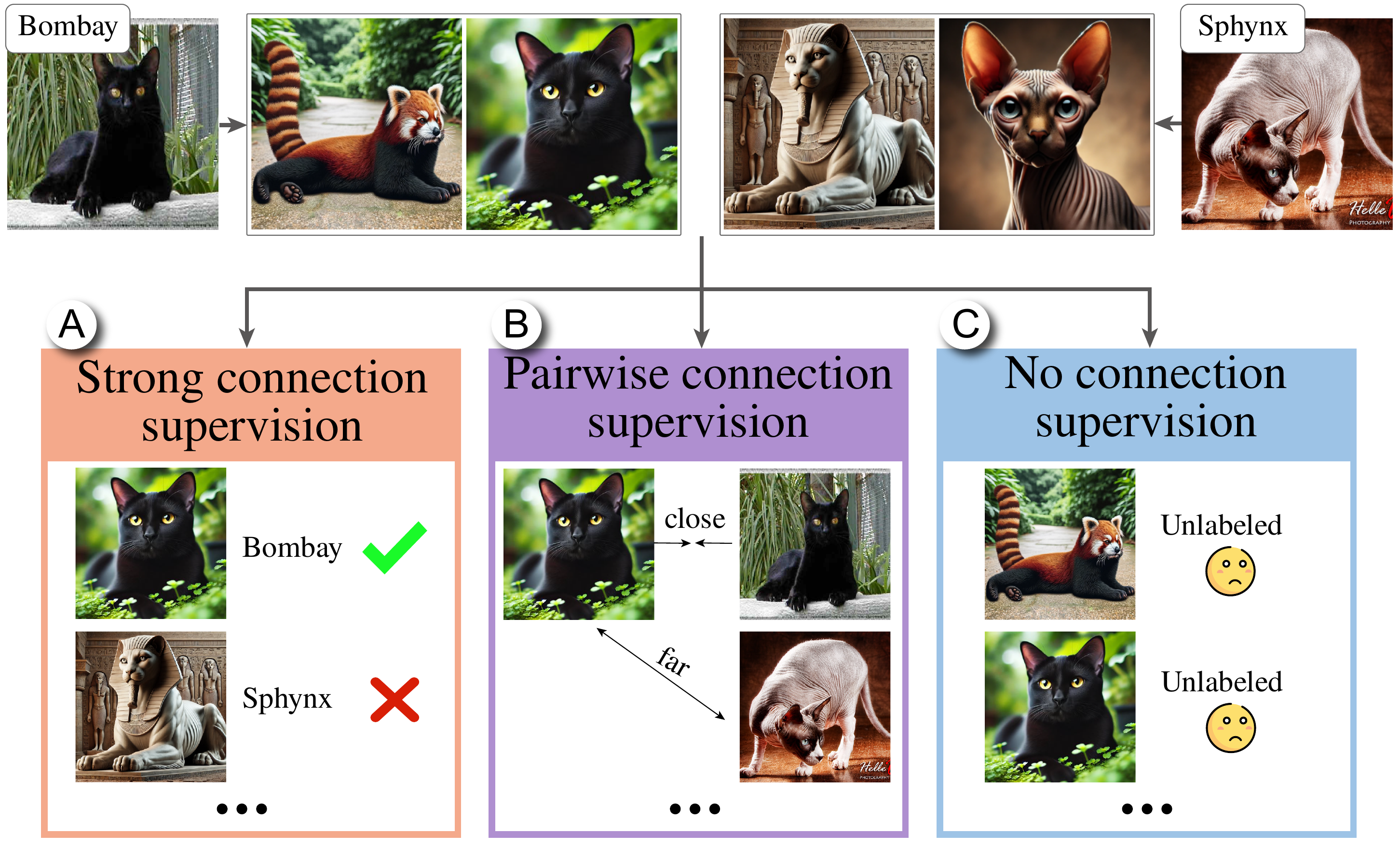}
    \caption{Three types of supervision for generated images: strong connection supervision, pairwise connection supervision, and no connection supervision.}
    \Description{}
    \label{fig:assumption}
\end{figure}

To address the negative effect caused by the noisy generated images, an effective strategy is re-weighting~\cite{yang2024interactive, zhang2024adapt}.
This strategy incorporates the supervision information of the \xiangting{original} images to adjust their weights in the training process, enabling the model to focus more on high-quality generated images while reducing attention to noisy ones.
According to the characteristics of generated images and our literature reviews~\cite{chen2023on, feng2023can, guo2020safe}, 
there are three types of supervision that can be used in re-weighting:
\begin{itemize}
\item \noindent{\emph{Strong connection supervision,}} which assumes each generated image belongs to one of the classes of interest and retains the same class as its original image (Fig.~\ref{fig:assumption}A).
All the existing methods only use this type of supervision for training downstream models.
\item \noindent{\emph{Pairwise connection supervision,}} which assumes a generated image is closer to its original image than the others (Fig.~\ref{fig:assumption}B).
\item \noindent{\emph{No connection supervision,}} where each generated image is treated as unlabeled (Fig.~\ref{fig:assumption}C).
\end{itemize}
These forms of supervision motivate us to explore the following question:
\textbf{are all these types of supervision helpful for the training process of the re-weighting?}

In this paper, we first theoretically investigate the effect of these three types of supervision.
For the \emph{strong connection supervision}, if the generated images are not of interest, their labels would be incorrect (\eg, the Great Sphynx of Giza in Fig.~\ref{fig:assumption}A).
We theoretically prove that such incorrect labels increase the upper bound of generalization risk by $\gamma O(\log c)$, where $c$ is the number of classes, and $\gamma$ is the percentage of the incorrect labels.
This result implies that strong connection supervision will hurt the training process.
For the \emph{pairwise connection supervision}, 
Lei~\etal~\cite{lei2023generalization} prove that the generalization performance can be guaranteed, even in the presence of low noise, where some generated images may not be closer to their original counterparts than others.
For the \emph{no connection supervision}, all generated images are treated as unlabeled ones.
The previous study~\cite{guo2020safe} theoretically proves that unlabeled images can always boost the training process, even if some of them belong to classes that are not of interest.

The theoretical results above show that the strong connection supervision hurts the training process, while the performance of the pairwise connection supervision and the no connection supervision can be guaranteed.
\textbf{This implies that the strong connection supervision used by the existing generative data augmentation methods is not the optimal choice.}
To this end, we develop {\sys}, a triplet-connection-based sample re-weighting method.
{\sys} leverages the pairwise connection supervision and no connection supervision through triplet loss and consistency regularization loss to improve model performance.
Theoretically, {\sys} can be integrated with any generative data augmentation methods and will never downgrade their performance.
Moreover, its generalization risk approaches the optimal in the order $O(\sqrt{d\ln (n)/n})$.
We experimentally validate the correctness of the theoretical analysis of the three types of supervision.
The experiments also show that our method outperforms the existing SOTA methods by $7.9\%$ on average over six natural image datasets and by $3.4\%$ on average over three medical datasets.
We further validate that our method can enhance the performance of different generative data augmentation methods.

In summary, the main contributions of our work are:

\begin{itemize}
    \item \textbf{Theoretical analysis} of three types of supervision for generated images, whose correctness is validated by the experiments.
    \item \textbf{A triplet-connection-based sample re-weighting meth-od} with a theoretical guarantee, which achieves SOTA classification performance.
    \item \textbf{A series of experimental results} demonstrate the effectiveness and efficiency of our method.
\end{itemize}
\section{Related Work}
Since our method focuses on leveraging image generative models-\cite{zhong2025generative} for data augmentation rather than enhancing the image generative models themselves, we focus solely on reviewing research related to data augmentation.
Data augmentation can be classified into two categories based on their evolution stages: rule-based data augmentation and generative data augmentation.

\myparagraph{Rule-based data augmentation} utilizes pre-defined rules to transform training data for improving generalization.
Such pre-defined rules include Random Erasing~\cite{devries2017improved, zhong2020random}, MixUp~\cite{hendrycks2019augmix, zhang2020does}, Transformation Selection~\cite{cubuk2019autoaugment, cubuk2020randaugment}, CutOut~\cite{devries2017improved} and CutMix~\cite{yun2019cutmix}. 
Although these rules are effective in certain scenarios, they cannot generate highly diversified images because they only vary images at the pixel level.
Moreover, as these methods employ random rules to transform images, they may introduce noisy data and lead to performance degradation~\cite{chen2024enhancing}.
To tackle these two issues, generative data augmentation has been proposed recently.

\myparagraph{Generative data augmentation} leverages pre-trained generative models~\cite{ho2020denoising, saharia2022photorealistic}, to generate diversified images to boost model performance. 
It can be further classified into perturbation-based~\cite{zhang2024expanding, antoniou2017data, samuel2024generating, zhu2024distribution, islam2024diffusemix} and prompt-based generative data augmentation~\cite{li2024semantic, trabucco2023effective, he2022synthetic}.
Perturbation-based generative data augmentation focuses on perturbing the latent space of generative models to generate diversified images.
\xiangting{The pioneering work along this line was proposed by Antoniou~\etal~\cite{antoniou2017data}.
They randomly added Gaussian noise to images and found that these images improved classification performance.}
The following studies optimize the perturbation process by adding different constraints, such as semantic-consistent constraint~\cite{zhang2024expanding}, concept-consistent constraint~\cite{samuel2024generating}, distribution-aware constraint~\cite{zhu2024distribution} and label-consistent constraint~\cite{islam2024diffusemix}.
Prompt-based generative data augmentation focuses on refining prompts to improve the quality of generated images.
%
Li~\etal~\cite{li2024semantic} utilized BLIP~\cite{li2022blip} to extract captions from images.
These captions were integrated into the prompts to maintain semantic consistency between the original and generated images. 
\xiangting{He~\etal~\cite{he2022synthetic} employed the T5 model~\cite{raffel2020exploring} to create a wider variety of prompts, thereby producing more diverse images.}
Since BLIP and T5 are not always perfect, these methods~\cite{li2024semantic, he2022synthetic} sometimes miss critical details, such as class names. 
To address this issue, Trabucco~\etal~\cite{trabucco2023effective} used Textual Inversion~\cite{galimage} to encode the information of images into several tokens and combined them with the prompts for generative models, effectively capturing detailed visual concepts.

%

%

Orthogonal to these methods, our work focuses on mitigating the negative effects of noisy images generated by these methods, thereby improving model performance.
For this purpose, we develop {\sys} to assign low weights to noisy images.
Since this method is agnostic to the generative data augmentation methods, it can be integrated with any generative data augmentation methods to improve model performance.
\section{Problem Formulation}
\label{sec:formatting}

In this section, we first briefly introduce the re-weighting problem for generative data augmentation. 
In this problem, we are given $n$ original images and their labels $\mathcal{D}_o = \left\{ (x_1, y_1), \dots, (x_n, y_n) \right\}$, and a generative data augmentation method $G$, such as GIF.
Usually, $n$ is small, $ x \in \mathcal{X} \in \mathbb{R}^D$, $y \in \mathcal{Y} = \{1, \dots, c\}$.
Here, $D$ is the number of input dimensions, and $c$ is the number of classes in the original data. 
For an original image $x_i$, the generative data augmentation method $G$ is applied to get $m$ generated images.
\begin{equation}
    \mathcal{D}_g^i = \{x_i^j = G(x_i) \}_{j=1}^{m},
\end{equation}
where $x_i^j$ is $j$-th image generated from $x_i$, \xiangting{$m$ is expansion ratio}.
Then we get a generated dataset $\mathcal{D}_g = \mathcal{D}_g^1 \cup \dots \cup \mathcal{D}_g^n$.

Given $\mathcal{D}_o$ and $\mathcal{D}_g$, the re-weighting problem for generative data augmentation is to learn a weight function $w: \mathbb{R}^D \to \mathbb{R}$ parameterized by $\alpha \in \mathbb{B}^d$ that maps a generated image to a weight value, and a classification model $h(x;\theta):\{\mathcal{X},\Theta\} \rightarrow \mathcal{Y}$ parameterized by $\theta$ to minimize the empirical risk on the clean original images:
\begin{align}
    &\min_{\alpha \in \mathbb{B}^d} \sum_{i=1}^n \ell ( h(x_i; \hat{\theta}), y_i) \\
    &\mathrm{s.t.} \ \hat{\theta} = \min_{\theta \in \Theta} \sum_{i=1}^{n} \ell(h(x_i; \theta), y_i)  + \sum_{i=1}^{n} \sum_{j=1}^{m} w(x_i^j; \alpha) \mathcal{L}(x_i^j, x_i; \theta),
    \label{equ:total} \notag
\end{align}
where $\ell$ refers to a loss function for classification, such as cross-entropy loss.

$\mathcal{L}(x_i^j, x_i; \theta)$ is the loss function for different types of supervision.
For strong connection supervision, $\mathcal{L}(x_i^j, x_i; \theta)= \ell(h(x_i^j; \theta), y_i)$.
Here, the label $y_i$ of the original image $x_i$ is assigned to the generated image $x_i^j$ and is used in the loss function $\ell$ for supervised learning.
For pairwise connection supervision, $\mathcal{L}(x_i^j, x_i; \theta) = \Phi(x_i^j, x_i; \theta)$ is a pairwise loss function, such as the triplet loss or the contrastive loss.
For no connection supervision, $\mathcal{L}(x_i^j, x_i; \theta)=\Omega(x_i^j; \theta)$, where $\Omega(x_i^j; \theta)$ is a regularization loss, such as the consistency regularization loss~\cite{berthelot2019mixmatch}.

\section{The {\sys} Framework}
To develop an effective re-weighting method, we first theoretically analyze the three types of supervision: strong connection supervision, pairwise connection supervision, and no connection supervision (Sec.~\ref{subsec:theory}).
Based on the theoretical analysis, we develop the {\sys} framework to dynamically assign low weights to noisy images and prove that it will never downgrade model performance and its generalization approaches the optimal in the order $O(\sqrt{d\ln (n)/n})$ (Sec.~\ref{subsec:formulation}).
Finally, we describe an online optimization method for the framework and show that this optimization method converges along with the iteration (Sec.~\ref{subsec:opt}).

\begin{figure*}
    \centering
    \includegraphics[width=.9\textwidth]{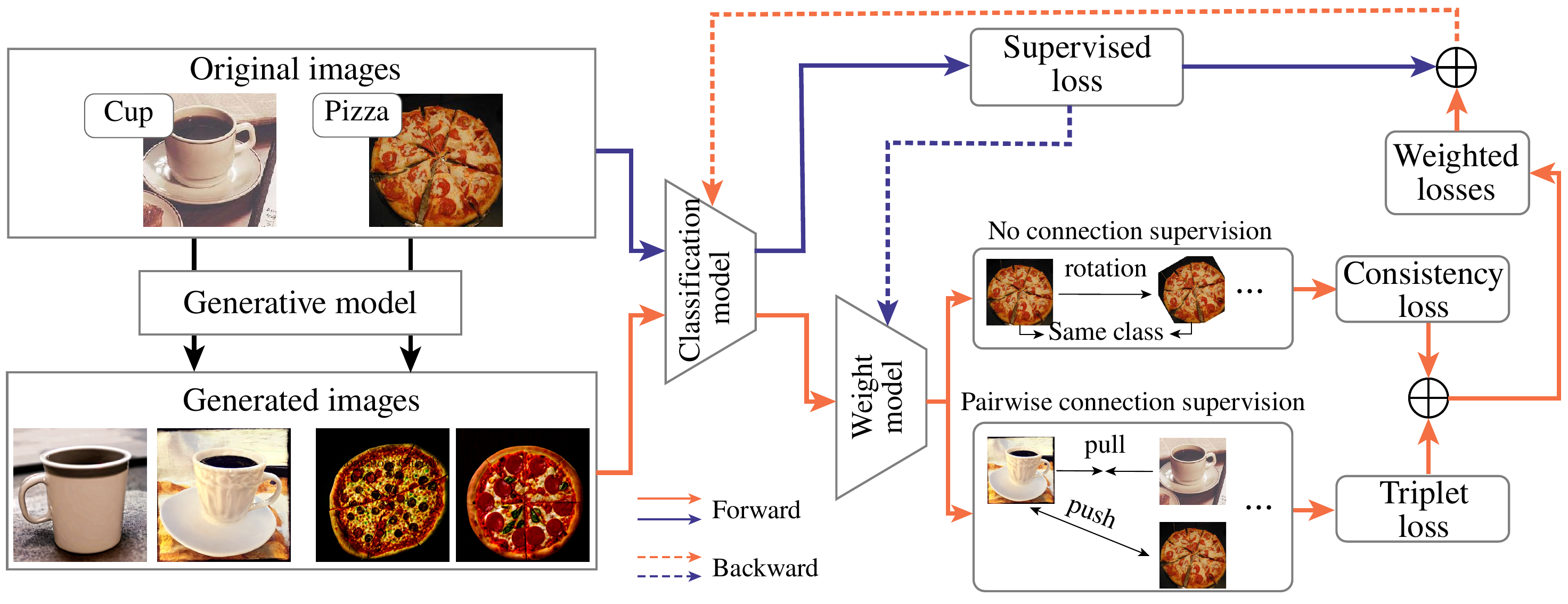}
    \caption{The framework of {\sys}.}
    \Description{}
    \label{fig:framework}
\end{figure*}

\subsection{Theoretical Analysis}
\label{subsec:theory}

To identify which of these three types of supervision are helpful for the training process, we first theoretically analyze their performance under the setting of generative data augmentation. 
\xiangting{All the proofs related to these theoretical analyses are provided in Appendix A.}

\myparagraph{Strong connection supervision}.
Under the strong connection supervision, each generated image is assigned the same class label as its associated original image.
However, some noisy images that are not of interest may be generated (\eg, Fig.~\ref{fig:samples}).
We group these noisy images into an extra class, the $(c+1)$-th class.
As we are only interested in the first $c$ classes, the classification model will never make predictions on this extra class.
Therefore, the classification results on these noisy images are always incorrect.
Let the percentage of such noisy images be denoted as $\gamma$ ($0<\gamma<1$).
Given a classification model $h$ parameterized by $\theta$ and a classification loss function $\ell$, the generalization risk of $h$ under clean images is defined as $R(\theta) = \mathbb{E}_{p(x,y)}[\ell (h(x;\theta), y)]$ and the generalization risk under the noisy images is defined as $R_{\mathcal{S}}(\theta)=\mathbb{E}_{p(x, \tilde{y})}[\ell(h(x;\theta), \tilde{y})]$.
Here, $p(x, y)$ and $p(x, \tilde{y})$ are the joint distributions of clean and noisy images, respectively.

\begin{theorem}
\label{theorem:strong}
Let $\tilde{\theta}$ and $\theta^*$ be the global minimizers of $R_{\mathcal{S}}(\theta)$ and $R(\theta)$ respectively, then
\begin{equation}
    0 < R(\tilde{\theta}) - R(\theta^*) \leq \gamma A,
\end{equation}
where $A$ is the maximum value of the loss function $\ell$.
\end{theorem}

Theorem~\ref{theorem:strong} reveals that with the presence of noisy generated images, the generalization risk will increase. 
If the classifier assigns equal probabilities to the $c$ classes for the noisy images, we have $R(\tilde{\theta}) - R(\theta^*) = \gamma \log c$.
This means that the generalization risk will never reach the optimal \xiangting{if $\gamma$ is large}.

\myparagraph{Pairwise connection supervision}.
Under the pairwise connection supervision, a generated image is assumed to be closer to its associated original image than the others.
This assumption is reasonable because generative data augmentation methods generate similar but distinct images to ensure relevance. 
To validate this, we also experimentally compared the generated images with their original ones and the others in the Pets dataset~\cite{parkhi2012pets}.
The result shows that over $92\%$ generated samples satisfy this assumption, which means the pairwise connection supervision is with low noise.
According to Theorem 4.13 and Lemma 5.8 in the work of Lei~\etal~\cite{lei2023generalization}, the generalization performance of a classification model trained on the representation after pairwise learning can be guaranteed, even in the presence of low noise.

\myparagraph{No connection supervision}.
Under the no connection supervision, all generated images are regarded as unlabeled, and some of them are not of interest.
According to the work of Guo~\etal~\cite{guo2020safe}, we have the following theorem.
\begin{theorem}[\textbf{Guo~\etal~\cite{guo2020safe}}]
\label{theorem:no}
Let $\theta^{SL}$ be the supervised model trained merely on the original images, \ie, 

$$\theta^{SL}=\arg min \sum_{i=1}^{n}\ell(h(x_i;\theta), y_i).$$

Define the empirical risk of the re-weighting model as:

\begin{equation}
    \hat{R}_{\mathcal{N}}(\theta)=\frac{1}{n}\sum_{i=1}^{n}\ell(h(x_i;\theta), y_i).
\end{equation}
Then we have the empirical risk of $\hat{\theta}$ returned by the re-weighting model to be never worse than $\theta^{SL}$:
\begin{equation}
    \hat{R}_{\mathcal{N}}(\hat{\theta}) \leq \hat{R}_{\mathcal{N}}(\theta^{SL}).
\end{equation}
\end{theorem}
Theorem~\ref{theorem:no} demonstrates that incorporating the no connection supervision during the training process will not negatively impact the model performance.

\subsection{Framework Formulation}
\label{subsec:formulation}
Based on the theoretical analysis above, we concluded that strong connection supervision impairs the training process, whereas both pairwise connection supervision and no connection supervision are beneficial.
Accordingly, we incorporate the pairwise connection supervision and the no connection supervision and develop the {\sys} framework (Fig.~\ref{fig:framework}), which optimizes the following objective:
\begin{equation}
\begin{aligned}
    &\min_{\alpha \in \mathbb{B}^d} \sum_{i=1}^n \ell (h(x_i; \hat{\theta}), y_i) \\
     &\mathrm{s.t.} \ \hat{\theta} = \min_{\theta \in \Theta} \sum_{i=1}^{n} \ell(h(x_i; \theta), y_i) \\
    & + \sum_{i=1}^{n} \sum_{j=1}^{m} w(x_i^j, x_i; \alpha) \cdot [\beta \Phi(x_i^j, x_i;\theta) + (1-\beta)\Omega(x_i^j;\theta)].
\end{aligned}
\label{equ:bi-level} 
\end{equation}

$\Phi(x_i^j, x_i; \theta)$ is the loss for pairwise connection supervision. 
In this framework, we choose the widely used triplet loss:
\begin{equation}
\begin{aligned}
&\Phi(x_i^j, x_i; \theta) = max(d(h(x_i^j; \theta), h(x_i; \theta))\\
&\quad\quad\quad\quad\quad-d(h(x_i^j; \theta), h(x_k; \theta))+\delta_{\mathrm{margin}}, 0).
\end{aligned}
\end{equation}
Here, $x_k$ is an original image randomly sampled from a class other than the class of $x_i$.
We do not use the more popular contrastive loss because it requires a very large amount of training data~\cite{chen2020simple}, but our framework focuses more on small-scale datasets.

$\Omega(x_i^j;\theta)$ is the loss for no connection supervision.
The consistency regularization loss is utilized here because of its simpleness and effectiveness:
\begin{equation}
\Omega(x_i^j;\theta) = {\|h(perturb(x_i^j); \theta) - h(x_i^j; \theta)\|_2^2,}
\end{equation}
where $perturb(x)$ is a stochastic operation, such as data rotation, data shearing, and model dropout.

$\beta$ is the term weight to balance $\Phi(x_i^j, x_i; \theta)$ and $\Omega(x_i^j;\theta)$.

As defined in Sec.~\ref{subsec:theory}, $\theta^{SL}$ denotes the supervised model trained merely on the original images, and $\hat{\theta}$ denotes our re-weighting model trained on both the original images and \textbf{weighted} generated images.
We further define $\theta^{NSL}$ as the supervised model trained on both the original and \textbf{unweighted} generated images (which is used for existing generative data augmentation methods, such as GIF~\cite{zhang2024expanding}):

\begin{equation}
\begin{aligned}
\theta^{NSL}=\arg \min_{\theta \in \Theta} &\sum_{i=1}^{n} \ell(h(x_i;\theta), y_i) + \sum_{i=1}^{n}\sum_{j=1}^{m}\ell(h(x_i^{j}; \theta), y_i).
\end{aligned}
\end{equation}

Then, we can prove that our method is never worse than the supervised model trained merely on the original images (\ie, $\theta^{SL}$), and our method will never downgrade the generative data augmentation methods (\ie, $\theta^{NSL}$).

\begin{theorem}
\label{theorem:guarantee}
Define the empirical risk of the re-weighting model as:
\begin{equation}
    \hat{R}(\theta)=\frac{1}{n}\sum_{i=1}^{n}\ell(h(x_i;\theta), y_i).
\end{equation}
Then we have:
\begin{equation}
\hat{R}(\hat{\theta}) \leq \hat{R}(\theta^{SL}), \hat{R}(\hat{\theta}) \leq \hat{R}(\theta^{NSL}).
\end{equation}
\end{theorem}

Moreover, we can prove that its generalization risk approaches the optimal in the order $O(\sqrt{d\ln (n)/n})$.

\begin{theorem} Assume that the loss function is $\lambda{-Lipschitz}$ continuous w.r.t. $\alpha$. 
Let $\alpha \in \mathbb{B}^d$ be the parameter of the example weight function $w$ in a d-dimensional unit ball. 
Let $n$ be the labeled data size. Define the generation risk as:
\begin{equation}
    R(\theta) = \mathbb{E}_{p(x,y)}[\ell(h(x; \theta), y)].
\end{equation}

Let ${\alpha}^* = \arg\max_{\alpha \in \mathbb{B}^d}R(\hat{\theta}(\alpha))$ be the optimal parameter in the unit ball, and $\hat{\alpha} = \arg\max_{\alpha \in \mathcal{A}}\hat{R}(\hat{\theta}(\alpha))$  be the empirically optima among a candidate set $\mathcal{A}$. 
With probability at least $1-\delta$ we have,
\begin{equation}
    R(\hat{\theta}(\alpha^*)) \leq R(\hat{\theta}(\hat{\alpha})) +  O(\sqrt{d\ln (n)/n}).
\end{equation}
\end{theorem}

\subsection{Optimization and Convergence Analysis}
\label{subsec:opt}
Eq.~(\ref{equ:bi-level}) is a bi-level optimization problem~\cite{bard2013practical}.
Directly optimizing it is time-consuming because there is no closed-form solution for $\theta$, and obtaining the optimal weights requires two nested loops of optimization.
To tackle this issue, we follow the work of Ren~\etal~\cite{ren2018learning} and utilize an online approximation method to solve it.

\begin{algorithm}[t]
    \caption{The {\sys} Framework.}
    \label{alg:TriWNET}
    \SetAlgoLined
    \KwIn{Original Dataset $\mathcal{D}_o$, generative data augmentation method $G$, max iterations $T$, generation ratio $m$.}
    \KwOut{Learned weight function parameter $\alpha_{T}$ and classification model parameter $\theta_{T}$.}
    \For{each image $x_i$ in $\mathcal{D}_o$} {
        
        \For{$j=1 \to m$} {
            $x_i^j = G(x_i)$\;
            Generated set $\mathcal{D}_g = \mathcal{D}_g \cup \{x_i^j\}$\;
        }
        
    }
    
    \For{$t = 0 \to T-1$} {
        $\{x, y\} \leftarrow SampleBatchOriginalData(\mathcal{D}_o)$\;        
        $\{x, y\} \leftarrow SampleBatchGeneratedData(\mathcal{D}_g)$\;
        Construct triplet set for $D_g$: $\{x_i^j, x_i, x_k\}$\;
        Compute loss of classification model: $\mathcal{L}^{cls}({\theta}_t, {\alpha}_t)$\;
        Update classification model: ${\theta}_{t+1} = {\theta}_{t} - \eta_{\theta} \cdot \nabla_{\theta}\mathcal{L}^{cls}({\theta}_t, {\alpha}_t)$\;
        Compute loss of weight function: $\mathcal{L}^{weight}({\theta}_{t+1})$\;
        Update weight function: ${\alpha}_{t+1} = {\alpha}_{t} - \eta_{\alpha} \cdot \nabla_{\alpha}\mathcal{L}^{weight}({\theta}_{t+1})$.
    }
\end{algorithm}

We denote the inner-level optimization as $\mathcal{L}^{cls}(\theta, \alpha)$ and the outer-level one as $\mathcal{L}^{weight}(\theta)$. 
The online approximation method alternatively updates $\theta$ and $\alpha$ to optimize Eq.~(\ref{equ:bi-level}). 
The complete algorithm is summarized in Algorithm~\ref{alg:TriWNET}.  

\myparagraph{Updating $\theta$ to optimize $\mathcal{L}^{cls}(\theta, \alpha)$}. 
Given the fixed parameter $\alpha_t$ of the weight function, we utilize gradient descent methods (\eg, SGD) to optimize the inner-level optimization:
\begin{equation}
    {\theta}_{t+1} = {\theta}_{t} - {\eta}_{\theta} {\nabla}_{\theta} \mathcal{L}^{cls}(\theta_t, \alpha_t).
\end{equation}

\myparagraph{Updating $\alpha$ to optimize $\mathcal{L}^{weight}(\theta)$}.
Given the updated $\theta_{t+1}$ of the classification model, we also utilize  gradient descent methods to optimize the outer-level optimization:
\begin{equation}
    {\alpha}_{t+1} = {\alpha}_t - {\eta}_{\alpha} {\nabla}_{\alpha} \mathcal{L}^{weight}({\theta}_{t+1}).
\end{equation}
Here, ${\eta}_{\theta}$ and ${\eta}_{\alpha}$ are the learning rates.

According to the work of Ren~\etal~\cite{ren2018learning} and Guo~\etal~\cite{guo2020safe}, this optimization method converges along with the iteration:

\begin{theorem} Suppose the supervised loss function is Lipschitz-smooth with constant $L \leq 2$, and the used losses have $\rho$-bounded gradients, then, by following the optimization algorithm, the labeled loss always monotonically decreases along with the iteration t, \ie,
\begin{equation}
    {\mathcal{L}}^{weight}({\theta}_{t+1}) \leq {\mathcal{L}}^{weight}({\theta}_{t}).
    \label{equ:con}
\end{equation}
Moreover, the equality in Eq.~(\ref{equ:con}) holds only when the gradient of the outer objective with respect to $\alpha$ becomes 0 at some iteration t, \ie,
\begin{equation}
    {\mathcal{L}}^{weight}({\theta}_{t+1}) = {\mathcal{L}}^{weight}({\theta}_{t}),
\end{equation}
if and only if
\begin{equation}
    {\nabla}_{\alpha} \mathcal{L}^{weight}(\theta_t) = 0.
\end{equation}
\end{theorem}
\section{Experiments}
\label{sec:exp}

\subsection{Experimental Settings}

\myparagraph{Datasets.} 
We evaluated the effectiveness and efficiency of our {\sys} method in image classification tasks over nine public datasets, consisting of both natural and medical datasets. 
The natural image datasets include Caltech~101~\cite{li2004caltech} and CIFAR100-Subset (CIFAR100-S)~\cite{krizhevsky2009cifar100} for coarse-grained classification, Cars~\cite{krause2013cars}, Flowers~\cite{nilsback2008flowers}, and Pets~\cite{parkhi2012pets} for fine-grained classification, and DTD~\cite{Cimpoi2014dtd} for texture classification. 
The medical image datasets~\cite{yang2021medical} cover a wide range of medical image types, including PathMNIST~\cite{kather2019pathmnist} for colon pathology, BreastMNIST~\cite{walid2020breastmnist} for breast ultrasound, and OrganSMNIST~\cite{xu2019organsmnist} for abdominal CT. 
A detailed summary of the statistics for datasets is provided in Appendix B.


\myparagraph{Baseline.} 
Theoretically, our {\sys} method can be used to enhance the performance of any generative data augmentation method. 
In this experiment, we chose GIF, a state-of-the-art (SOTA) generative data augmentation method, as a representative example. 
The effectiveness of our method was validated by comparing the performance of GIF with and without our method.
In addition, we also compared them with representative rule-based data augmentation methods, including  CutOut~\cite{devries2017improved},
GridMask~\cite{chen2020gridmask},
RandAugment~\cite{cubuk2020randaugment}, TrivialAugment~\cite{muller2021trivialaugment}, TeachAugment~\cite{suzuki2022teachaugment}, MADAug~\cite{hou2023learn} and EntAugment~\cite{yang2024entaugment}.
The performance was evaluated by measuring the classification accuracy on the test dataset.







\myparagraph{Implementation details}.
For GIF~\cite{zhang2024expanding} with our method, we utilized \textit{Stable Diffusion} v.1.4~\cite{rombach2022high} for image generation. The hyper-parameters of GIF, such as the strength of Stable Diffusion, were set in the same way as the original GIF paper.
After the data augmentation, both the original and the generated images were used to train a ResNet-50 model~\cite{he2016deep} from scratch.
Similar to GIF, we limited the training process to a maximum of 100,000 iterations.

\begin{table}[b]
    \centering
    \caption{Accuracy (in $\%$) comparison of different combinations of the three types of supervision on Pets.}
        \begin{tabular}{cccc}
            \toprule
             Strong & Pairwise & No & \multirow{2}{*}{Accuracy} \\
              connection & connection & connection & \\
            \midrule
            
             \checkmark &  &  & 75.3 \\ \midrule
              & \checkmark & &  78.3\\
             \checkmark & \checkmark & & 74.3 ($\downarrow$ 4.0) \\ \midrule
              &  & \checkmark & 75.7 \\
             \checkmark &  & \checkmark & 74.5 ($\downarrow$ 1.2) \\ \midrule
              & {\checkmark} & {\checkmark} & \textbf{82.3} \\
             \checkmark & \checkmark & \checkmark & 77.1 ($\downarrow$ 5.2) \\
 
            \bottomrule
            
        \end{tabular}
    \label{tab: ablation}
\end{table}

\begin{table*}[b]
    \centering
    \caption{Accuracy (in \%) of a ResNet-50 trained from scratch on original and generated images by different methods. }
    \resizebox{\textwidth}{!}{
    \begin{tabular}{lccccccccccc}
         \toprule
         \multirow{2}{*}{\textbf{Dataset}} & \multicolumn{7}{c}{\textbf{Natural image datasets}} & \multicolumn{4}{c}{\textbf{Medical image datasets}} \\
        \cmidrule(lr){2-8} \cmidrule(lr){9-12}
        & Caltech 101 & Cars & Flowers & DTD & CIFAR100-S & Pets & \textbf{Average} & PathMNIST & BreastMNIST & OrganSMNIST & \textbf{Average} \\
        \midrule
         original & 26.3 & 19.8 & 74.1 & 23.1 & 35.0 & 6.8 & 30.9 & 72.4 & 55.8 & 76.3 & 68.2\\
         CutOut & 51.5 & 25.8 & 77.8 & 24.2 & 44.3 & 38.7 & 43.7 & 78.8 & 66.7 & 78.3 & 74.6\\
         GridMask & 51.6 & 28.4 & 80.7 & 25.3 & 48.2 & 37.6 & 45.3 & 78.4 & 66.8 & 78.9 & 74.7 \\
         RandAugment & 57.8 & 43.2 & 83.8 & 28.7 & 46.7 & 48.0 & 51.4 & 79.2 & 68.7 & 79.6 & 75.8 \\
         TrivialAugment & 49.9 & 21.1 & 81.8 & 28.0 & 37.3 & 5.9 & 37.3 & 83.2 & 61.0 & 78.2 & 74.1 \\
         TeachAugment & 70.5 & 25.9 & 58.7 & 51.5 & 31.3 & 68.7 & 51.1 & 79.6 & 74.3 &  77.2 & 77.0 \\
         MADAug & 65.5 & 55.3 & 84.7 & 44.4 & 38.2 & 67.8 & 59.3 & 75.6 & 73.1 & 75.9 & 74.9 \\
         EntAugment & 70.7 & 69.0 & 75.9 & 40.8 & 52.4 & 67.8 & 62.8 & 86.2 & 74.4 & 77.7 & 79.4 \\
         \hdashline
         GIF-w/o-TriReWeight & 65.1 & 75.7 & 88.3 & 43.4 & 61.1 & 73.4 & 67.8 & 86.9 & 77.4 & 80.7 & 81.7\\
         GIF-w/-TriReWeight  & \textbf{80.5} & \textbf{77.5} & \textbf{88.9} & \textbf{59.1} & \textbf{62.8} & \textbf{85.3} & \textbf{75.7} (\textcolor{red}{+7.9}) & \textbf{88.7} & \textbf{85.3} & \textbf{81.2} & \textbf{85.1} (\textcolor{red}{+3.4})\\ 
         
         \bottomrule
    \end{tabular}
    }
    \label{tab:overall}
\end{table*}

\subsection{Validating the Correctness of the Theoretical Analysis (Ablation Study) }
We first validated the correctness of the theoretical analysis in Sec.~\ref{subsec:theory}.
This experiment was conducted on four datasets,
including both natural (\ie, Pets, DTD, and Flowers) and medical (\ie, OrganSMNIST) image datasets.
We evaluated the accuracy of different combinations of the three types of supervision.
Since the comparison of the four datasets leads to a similar conclusion, we only present the results on the $20 \times$-expanded Pets dataset in this manuscript and include the others in Appendix C.
As shown in Table~\ref{tab: ablation}, when the strong connection supervision is combined with either of the other two or with both of them, the accuracy dropped, ranging from $1.2\%$ to $5.2\%$.
However, when only pairwise connection supervision and no connection supervision were combined, the accuracy was improved compared to using either alone (+$4.0\%$ and +$6.6\%$), achieving the highest accuracy of $82.3\%$.
These experimental results are consistent with our theoretical analysis in Sec.~\ref{subsec:theory}. 

\subsection{Overall Performance}

\myparagraph{Effectiveness.} 
As shown in Table~\ref{tab:overall}, our method outperforms the rule-based data augmentation methods by a large margin.
Compared to GIF without our method, GIF with our method performs better in all nine datasets.
Overall, GIF with our method achieves an average performance improvement of 7.9\% across six natural image datasets and 3.4\% across three medical image datasets. 
This validates that not all images generated by GIF are correct and desired.
Incorporating these noisy images into the training process will harm model performance.
By assigning low weights to such noisy images, our method reduces their negative effect on the training process and thus improves the model performance.
This demonstrates the effectiveness of our re-weighting strategy.

\begin{figure*}[t]
  \centering
   \includegraphics[width=.85\textwidth]{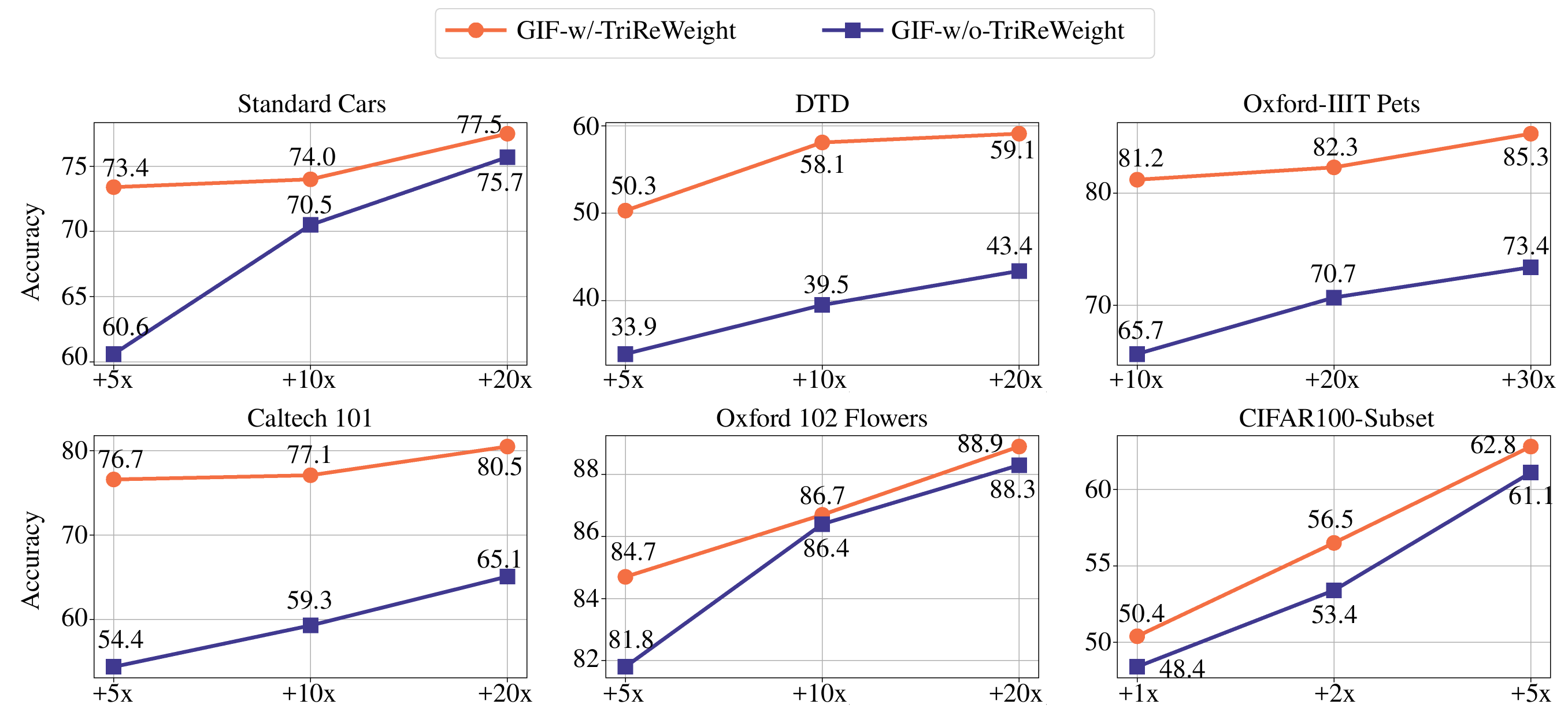}
   \caption{Accuracy (in \%) of a ResNet-50 trained from scratch on the original and generated dataset by GIF with and without our method with different expansion ratios.}
   \Description{}
   \label{fig:efficiency}
   \vspace{-0.5em}
\end{figure*}

We also compared GIF with and without our method under different expansion ratios.
The expansion ratios for different datasets were set in the same way as in GIF.
As shown in Fig.~\ref{fig:efficiency}, 
GIF with our method consistently requires fewer generated images compared to GIF without our method.
For the Pets, Caltech 101, and DTD datasets, a $5 \times$ expansion obtained with our method achieves higher accuracy than a $20 \times$ expansion obtained without our method.
This implies that our method achieves at least a $4 \times$ efficiency improvement on these datasets.

\begin{figure}[t]
    \centering
    \includegraphics[width=.8\linewidth]{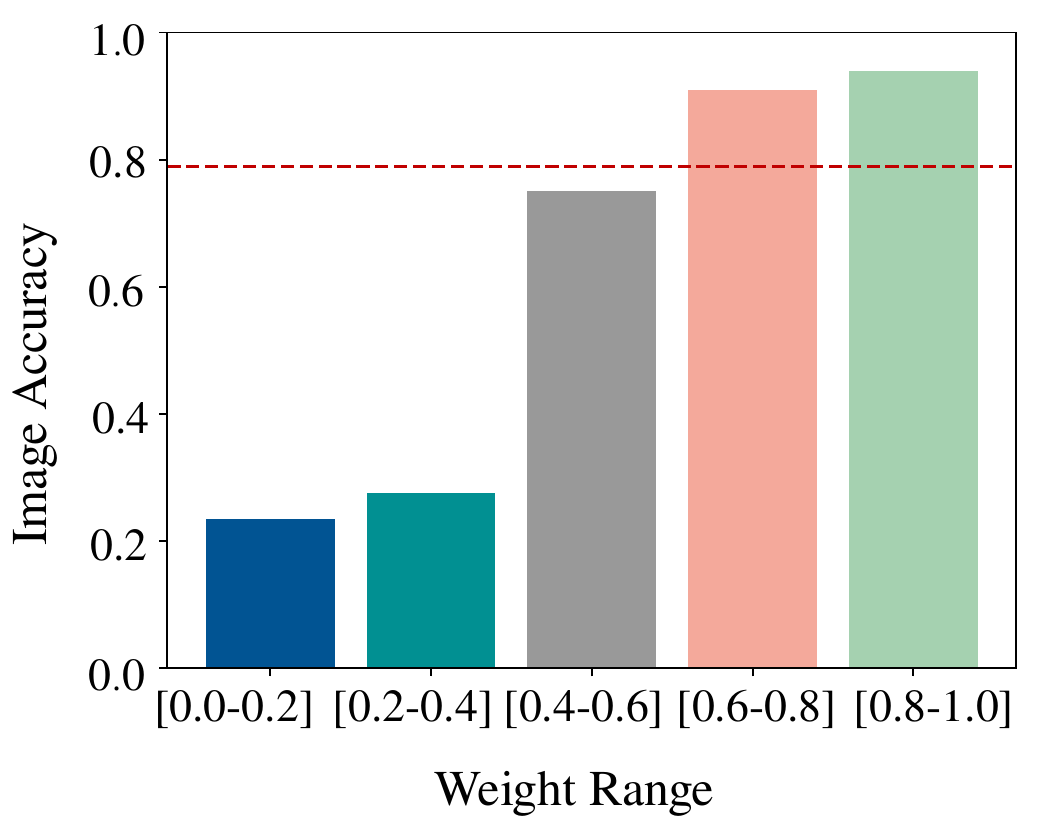}
    \caption{Image accuracy on Pets under different weights.}
    \Description{}
    \label{fig:image_accuracy}
\end{figure}

\begin{figure}[t]
  \centering
   \includegraphics[width=.8\linewidth]{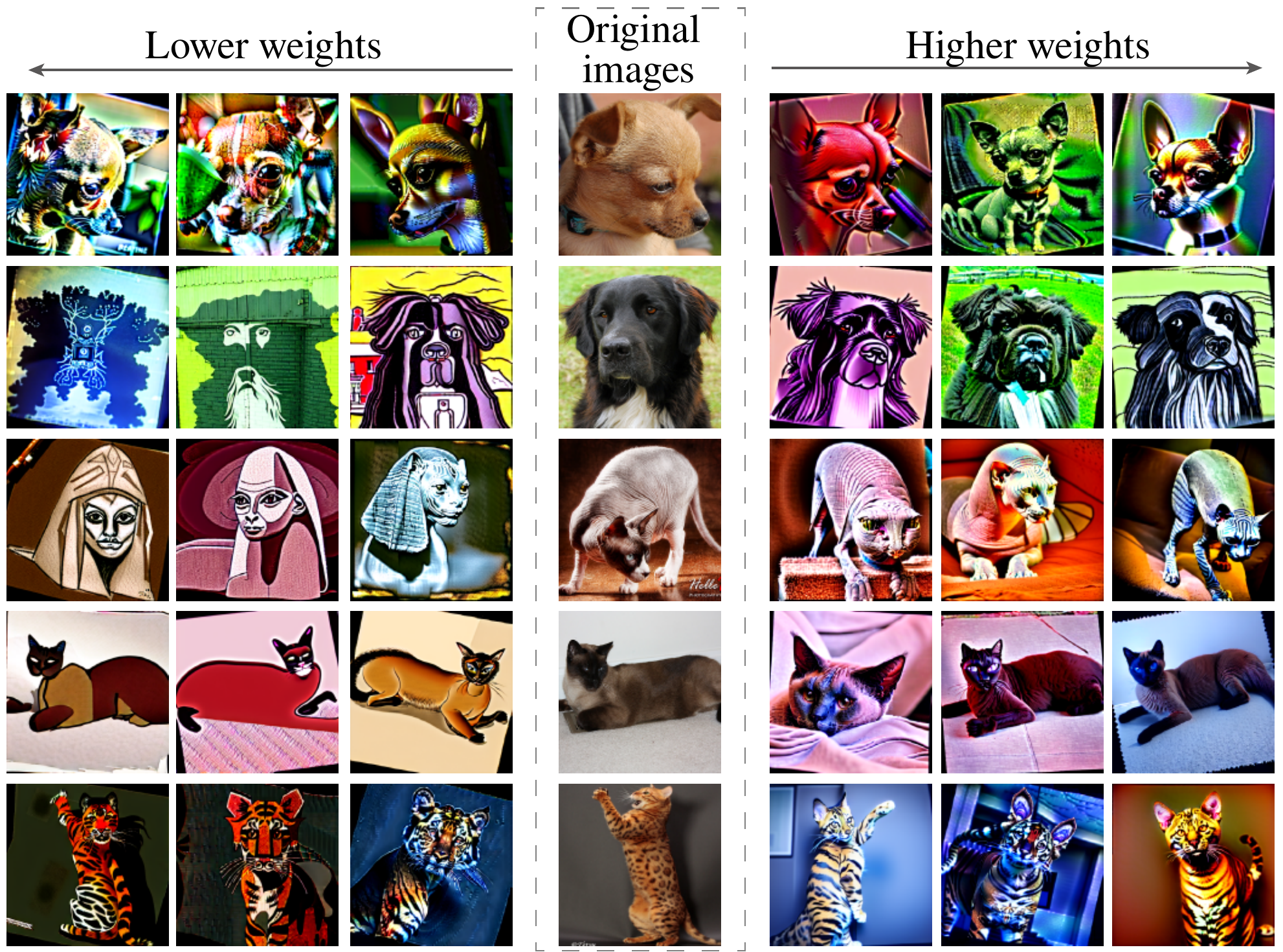}
   \caption{Exemplar original images on Pets and their associated generated images ordered by their weights.}
   \Description{}
   \label{fig:weight}
\end{figure}

\myparagraph{Computational efficiency for inference.}
Since the inference cost is typically higher than the training cost in real-world applications, we focus on comparing the inference cost.
Our method only re-weights the generated images during the training process without changing the inference structure of the model.
Therefore, it does not introduce any additional computational cost during inference.
To validate this, we randomly sampled $1,000$ images and ran a ResNet-50 trained by GIF with and without our method to compare the inference time.
The experiment was conducted on an Nvidia A100 GPU.
It takes 6.21 seconds for GIF without our method and 6.11 seconds for GIF with our method.


\myparagraph{Image accuracy improvement after re-weighting}.
To demonstrate that our method can effectively assign low weights to noisy generated images more intuitively, we calculated the accuracy of generated images across the six natural datasets.
We did not use the medical image datasets due to their low resolution, which made labeling difficult, even for doctors.
The calculation of image accuracy requires the ground truth labels of generated images, which are not available.
Therefore, we randomly sampled some generated images and manually labeled them.
More specifically, we divided the weight range into five equal intervals.
For each interval, we randomly sampled 200 generated images, resulting in a total of $1,000$ images.
To ensure the quality of the manual labels, we invited two trained annotators to label these generated images independently.
Then, for the images they had assigned different labels to, they discussed and worked to reach a consensus.

The image accuracy under different weight intervals for the Pets dataset is shown in Fig.~\ref{fig:image_accuracy}.
The results on the other datasets are shown in Appendix D.
The red line represents the average image accuracy of all generated images.
As we can see, as the weights increase, the image accuracy also improves.
When the weight is larger than average, the image accuracy is higher than the average.
The results show that our method can effectively assign low weights to noisy generated images.
In Fig.~\ref{fig:weight}, we also present several original images and their generated images ordered by their weights.
As we can see, the generated images become less relevant to the original ones when the weights decrease.
For example, in the third row in Fig.~\ref{fig:weight}, the generated images with higher weights are more like Sphynx cats as the original one, but the generated images with lower weights are more like ``Great Sphynx of Giza''.
In the fifth row, the generated images with higher weights are still like cats, but the generated images with lower weights are more like tigers, which is the class that is not of interest.

\subsection{Generalization}
To demonstrate the effectiveness of our method, we also conducted a set of experiments to show the generalization ability of our method to other SOTA generative data augmentation methods, other SOTA generative models, and various classification model backbones.

\myparagraph{Generalization to other SOTA generative data augmentation methods. }
\xiangting{Here, we conducted experiments to validate that our method can be used to enhance other generative data augmentation methods, including DiffuseMix~\cite{wu2023diffumask}, Real Guidance~\cite{he2022synthetic}, and Da-Fusion~\cite{trabucco2023effective}.}
We compared the performance of these methods with and without our method.
Since the images generated by DiffuseMix are used to train models from scratch, while the images generated by Real Guidance and Da-Fusion are used to fine-tune models pre-trained on ImageNet in their original papers, we compared these methods separately to ensure a fair comparison.

For DiffuseMix that trains models from scratch, the experiment was conducted on three fine-grained datasets, including Caltech-UCSD Birds-200-2011 (CUB)~\cite{wah2011caltech}, Aircraft~\cite{maji2013fine}, and Cars~\cite{krause20133d}, which were used in the original DiffuseMix paper.
Since the original paper did not specify the expansion ratios, we set it as $4 \times$ across all three datasets.
Thus, the results of DiffuseMix reported in this paper differ slightly from those in the original paper.
However, this did not affect the conclusion of this experiment because our proposed method focuses on better leveraging generated images.
If the generated images improve, our method will also perform better.
As shown in Table~\ref{tab:combine}, the accuracy of DiffuseMix with our method is higher than DiffuseMix without our method across all three datasets.
It achieves a gain of 2.89\% accuracy on average.

For Real Guidance and Da-Fusion that fine-tune a ResNet-50 pre-trained on ImageNet, 
\xiangting{we conducted experiments} on the PascalVOC~\cite{DBLP:journals/ijcv/EveringhamGWWZ10} and COCO~\cite{DBLP:conf/eccv/LinMBHPRDZ14} datasets. 
Following the settings in the original Da-Fusion paper, we used 4 original images per class and got $10 \times$-expanded datasets for model training.
As shown in Table~\ref{tab:combine-pretrain}, our method achieves 5.46\% and 2.18\% improvement compared to Real Guidance and Da-Fusion on average, respectively.

These results demonstrate that our method is not designed for a specific generative data augmentation method but can be used for enhancing other generative data augmentation methods.
Moreover, our method can enhance both models trained from scratch and those pre-trained on large-scale datasets, demonstrating the robustness and generalization ability of our method.
\begin{table}[t]
    \centering
    \caption{Accuracy (in \%) comparison between DiffuseMix with and without our method \xiangting{for training a ResNet-50 from scratch}.}
    \resizebox{\linewidth}{!}{
    \begin{tabular}{lcccc}
         \toprule
         Method & CUB & Aircraft & Cars & \textbf{Average} \\
         \midrule
         DiffuseMix-w/o-{\sys} & 66.21 & 79.69 & 89.35 & 78.42 \\
         DiffuseMix-w/-{\sys} & 69.35 & 83.20 & 91.37 & \textbf{81.31} (\textcolor{red}{+2.89}) \\
         \bottomrule
    \end{tabular}
    \label{tab:combine}
    }
\end{table}

\begin{table}[t]
    \centering
    \caption{\xiangting{Accuracy (in \%) comparison between Real Guidance and Da-Fusion with and without our method for fine-tuning a ResNet-50 pre-trained on ImageNet.}}
    \resizebox{\linewidth}{!}{
    \begin{tabular}{lccc}
         \toprule
         Method & Pascal & COCO & \textbf{Average} \\
         \midrule
         Real Guidance-w/o-{\sys} & 70.81 & 40.15 & 55.48 \\
         Da-Fusion-w/o-{\sys} & 74.23 & 47.87 & 61.05 \\
         \midrule
         Real Guidance-w/-{\sys} & 73.15 & 48.72 & \textbf{60.94} (\textcolor{red}{+5.46}) \\
         Da-Fusion-w/-{\sys} & 75.92 & 50.54 & \textbf{63.23} (\textcolor{red}{+2.18}) \\
         \bottomrule
    \end{tabular}
    \label{tab:combine-pretrain}
    }
\end{table}

\myparagraph{Generalization to other SOTA generative models. }
\xiangting{To further demonstrate that our method can be combined with other generative models, we get the $5 \times$-expanded datasets by various SOTA diffusion models, including SD~1.4~\cite{rombach2022high}, SD~2.1~\cite{ho2022classifier}, SD~3.5~\cite{esser2024scaling}, and SDXL~\cite{podell2023sdxl}, to train a ResNet-50 from scratch.}
Table~\ref{tab:car-model-size} shows the results on the Pets dataset, and the remaining results \xiangting{on additional 8 datasets} can be found in Appendix E.
Our method consistently improves performance with different SOTA diffusion models.
It validates that our method will remain effective even as more powerful generative models are developed in the future.

\begin{table}[t]
    \centering
    \caption{Accuracy (\%) with different SOTA diffusion models for training a ResNet-50 from scratch on Pets.}
    \resizebox{\linewidth}{!}{
    \begin{tabular}{l|ccc|c}
         \hline
         \multirow{2}{*}{SD Models} & SD 1.4 & SD 2.1 &  SD 3.5 & SDXL \\
          & (860M) & (865M) & (8.1B) & (2.6B) \\
         \hline
         w/o-TriReWeight & 41.4 & 42.4 & 62.1 & 68.2 \\
         w/-TriReWeight & \textbf{67.0} \footnotesize (\textcolor{red}{+25.6}) & \textbf{68.9} 
 \footnotesize(\textcolor{red}{+26.5}) &  \textbf{77.4} \footnotesize(\textcolor{red}{+15.3}) & \textbf{78.6} \footnotesize(\textcolor{red}{+10.4}) \\
         \hline
    \end{tabular}
    }
    \label{tab:car-model-size}
\end{table}

\myparagraph{Generalization to other classification model backbones.}
In the experiments above, we proved that our method can enhance both models trained from scratch and those pre-trained on large-scale datasets.
In this experiment, we further validate that our method can also enhance the performance with different classification model backbones.

We applied our method to different backbones, including ResNeXt-50~\cite{xie2017aggregated}, WideResNet-50~\cite{zagoruyko2016wide}, and MobileNet-v2~\cite{sandler2018mobilenetv2}, and trained them from scratch on $5 \times$-expanded datasets generated by GIF, with and without our method.
Table~\ref{Tab:pets_backbones} shows the results on the Pets dataset, and the remaining results can be found in Appendix F.
As shown in Table~\ref{Tab:pets_backbones}, GIF with our method consistently performs better than GIF without our method across different backbones. 

\begin{table}[t]
\centering
\caption{Model performance of various backbones trained on GIF with and without our method on Pets. 
}
\resizebox{\linewidth}{!}{
\begin{tabular}{lcccc}
    \toprule

    Models & ResNeXt-50 & WideResNet-50 & MobileNet-v2 & Average \\
    \midrule
    w/o-{\sys} & 56.5 & 70.9 & 60.6 & 62.7 \\
    w/-{\sys} & \textbf{64.5} & \textbf{71.3} & \textbf{69.5} & \textbf{68.4} (\textcolor{red}{+5.7}) \\
    \bottomrule
\end{tabular}
}
\label{Tab:pets_backbones}
\end{table}
\section{Conclusion}

This paper tackles the issue of performance degradation in generative data augmentation due to noisy generated images.
Drawing inspiration from sample re-weighting, we first theoretically analyze three types of supervision tailored to the generative data augmentation setting.
Based on the theoretical analysis, we introduced a training strategy {\sys}, which is orthogonal to other generative data augmentation methods.
Extensive experiments validate the correctness of the theoretical analysis and demonstrate that our proposal achieves SOTA accuracy. 

There are several directions for future research. For instance, extending our method to enhance other types of data, such as text, would be valuable. Additionally, exploring how to combine our method with rule-based data augmentation methods to further improve classification performance is an intriguing avenue for future investigation.

\begin{acks}
The work is supported by the National Natural Science Foundation of China (Grant Nos. 62225205, 62373141, 62402167), the Science and Technology Program of Changsha (kq2502272, kh2301011), the Science and Technology Innovation Program of Hunan Province (No. 2023ZJ1080), the Hunan Natural Science Foundation under the grant 2025JJ60419, the Major Science and Technology Research Projects of Hunan Province (Grant Nos. 2024QK2010, 2024QK2009), the Yunnan Provincial Major Science and Technology Special Plan Projects (No. 202502AD080009).

We would like to express our gratitude to Jinhui Zhao, an undergraduate student at Hunan University, for his valuable contributions to the experimental evaluation in this work.
\end{acks}

\bibliographystyle{ACM-Reference-Format}
\balance
\bibliography{main}

\newpage
\appendix
\section{Proofs}
\label{sup:proof}
\subsection{Proof of Theorem 4.1}
Given a classification model $h$ parameterized by $\theta$ and a classification loss function $\ell$, the generalization risk of $h$ under clean images is defined as  $R(\theta) = \mathbb{E}_{p(x,y)}[\ell (h(x;\theta), y)]$ and the generalization risk under the noisy images is defined as $R_{\mathcal{S}}(\theta)=\mathbb{E}_{p(x, \tilde{y})}[\ell(h(x;\theta), \tilde{y})]$.

\setcounter{theorem}{0} 
\renewcommand{\thetheorem}{4.\arabic{theorem}} 
\begin{theorem}\label{proof:1}
Let $\tilde{\theta}$ and $\theta^*$ be the global minimizers of $R_{\mathcal{S}}(\theta)$ and $R(\theta)$ respectively, then
\begin{equation}
    0 < R(\tilde{\theta}) - R(\theta^*) \leq \gamma A,
\end{equation}
where $A$ is the maximum value of the loss function $\ell$.
\end{theorem}

\textbf{Proof.} Under the generative data augmentation setting, we group all noisy images into an extra class, the ($c+1$)-th class. 
Let the percentage of such noisy images be denoted as $\gamma$ ($0<\gamma<1$), and such noisy images be denoted as $\tilde{x}$.
We assume the labels of the first $c$ classes are always correct, and the images of the ($c+1$)-th class are assigned the label of the first $c$ classes with an equal probability $\frac{1}{c}$, that is $p(\tilde{y}=i|y=i)=1 \ \mathrm{for }\ i=1,2,...,c$, $p(\tilde{y}=i|y=c+1)=\frac{1}{c}\ \mathrm{for }\ i=1,2,...,c$.
Then we have
\begin{equation}
\begin{aligned}
    &R_{\mathcal{S}}(\theta) \\
    &= \mathbb{E}_{p(x, \tilde{y})}[\ell(h(x;\theta), \tilde{y})]\\
    &= \mathbb{E}_x \mathbb{E}_{p(y|x)} \mathbb{E}_{p(\tilde{y}|y)}[\ell(h(x; \theta),\tilde{y})]\\
    &= \mathbb{E}_{p(x,y)}[\ell(h(x; \theta),y) + \frac{1}{c} \cdot \gamma \sum_{j}^{c}\ell(h(\tilde{x};\theta), j)]\\
    &= R(\theta) + \frac{\gamma}{c}(\sum_{j=1}^c \ell(h(\tilde{x};\theta), j))
\end{aligned}
\end{equation}
We assume the upper bound of $\ell$ is $A$. We have
\begin{equation}
    R_{\mathcal{S}}(\theta) \leq R(\theta) + \gamma A.
\end{equation}
Then we have
\begin{equation}
     0 \leq R_{\mathcal{S}}(\theta) - R(\theta) \leq \gamma A.
\end{equation}

\subsection{Proof of Theorem 4.2}
\label{proof:2}

\begin{theorem}
Let $\theta^{SL}$ be the supervised model trained merely on the original images, \ie, 
$$\theta^{SL}=\arg min \sum_{i=1}^{n}\ell(h(x_i;\theta), y_i).$$
Define the empirical risk of the re-weighting model as:
\begin{equation}
    \hat{R}_{\mathcal{N}}(\theta)=\frac{1}{n}\sum_{i=1}^{n}\ell(h(x_i;\theta), y_i).
\end{equation}
Then we have the empirical risk of $\hat{\theta}$ returned by the re-weighting model to be never worse than $\theta^{SL}$:
\begin{equation}
    \hat{R}_{\mathcal{N}}(\hat{\theta}) \leq \hat{R}_{\mathcal{N}}(\theta^{SL}).
\end{equation}
\end{theorem}
\textbf{Proof.} 
By setting the weights of all generated images to zero, optimizing $\hat{R}_{\mathcal{N}}(\hat{\theta})$ equals optimizing $\hat{R}_{\mathcal{N}}(\theta^{SL})$, and we have  $\hat{R}_{\mathcal{N}}(\hat{\theta}) = \hat{R}_{\mathcal{N}}(\theta^{SL})$.
Therefore, $\hat{R}_{\mathcal{N}}(\hat{\theta}) \leq \hat{R}_{\mathcal{N}}(\theta^{SL}).$


\subsection{Proof of Theorem 4.3}
\label{proof:3}
\begin{theorem}
Define the empirical risk of the re-weighting model as:
\begin{equation}
    \hat{R}(\theta)=\frac{1}{n}\sum_{i=1}^{n}\ell(h(x_i;\theta), y_i).
\end{equation}
Then we have:
\begin{equation}
\begin{aligned}
    & \hat{R}(\hat{\theta}) \leq \hat{R}(\theta^{SL}),\\
    & \hat{R}(\hat{\theta}) \leq \hat{R}(\theta^{NSL}).
\end{aligned}
\end{equation}
\end{theorem}
\textbf{Proof.} For $\hat{R}(\hat{\theta}) \leq \hat{R}(\theta^{SL})$, the proof of it is shown in Sec.~\ref{proof:3}.

For $\hat{R}(\hat{\theta}) \leq \hat{R}(\theta^{NSL})$, by setting the weights of all generated images to one, optimizing $\hat{R}(\hat{\theta})$ equals optimizing $\hat{R}(\theta^{NSL})$ and we have $\hat{R}(\hat{\theta}) = \hat{R}(\theta^{NSL})$. 
Therefore, $\hat{R}(\hat{\theta}) \leq \hat{R}(\theta^{NSL})$.

\subsection{Proof of Theorem 4.4}
\begin{theorem} Assume that the loss function is $\lambda{-Lipschitz}$ continuous w.r.t. $\alpha$. 
Let $\alpha \in \mathbb{B}^d$ be the parameter of example weighting function $w$ in a d-dimensional unit ball. 
Let $n$ be the labeled data size. Define the generation risk as:
\begin{equation}
    R(\theta) = \mathbb{E}_{p(x,y)}[\ell(h(x; \theta), y)].
\end{equation}
Let ${\alpha}^* = \arg\max_{\alpha \in \mathbb{B}^d}R(\hat{\theta}(\alpha))$ be the optimal parameter in the unit ball, and $\hat{\alpha} = \arg\max_{\alpha \in \mathcal{A}}\hat{R}(\hat{\theta}(\alpha))$  be the empirically optima among a candidate set $\mathcal{A}$. 
With probability at least $1-\delta$ we have,
\begin{equation}
    R(\hat{\theta}(\alpha^*)) \leq R(\hat{\theta}(\hat{\alpha})) +  O(\sqrt{d\ln (n)/n}).
\end{equation}
\end{theorem}
\textbf{Proof.} Guo~\etal~\cite{guo2020safe} have proved in their theorem 4:
\begin{equation}
    R(\hat{\theta}(\alpha^*)) \leq R(\hat{\theta}(\hat{\alpha})) + \frac{3\lambda + \sqrt{4d\ln{(n)} + 8\ln{(2/\delta)} }}{\sqrt{n}},
\end{equation}
which proves that $R(\hat{\theta}(\alpha^*)) - R(\hat{\theta}(\hat{\alpha}))$ decreases at a rate of\\ $O(\sqrt{d\ln (n)/n})$. 

Therefore, we have $ R(\hat{\theta}(\alpha^*)) \leq R(\hat{\theta}(\hat{\alpha})) +  O(\sqrt{d\ln (n)/n})$.

\subsection{Proof of Theorem 4.5}
\begin{theorem} Suppose the supervised loss function is Lipschitz-smooth with constant $L \leq 2$, and the used losses have $\rho$-bounded gradients, then, by following the optimization algorithm, the labeled loss always monotonically decreases along with the iteration t, \ie,
\begin{equation}
    {\mathcal{L}}^{weight}({\theta}_{t+1}) \leq {\mathcal{L}}^{weight}({\theta}_{t}).
    \label{equ:con_1}
\end{equation}
Furthermore, the equality in Eq.~(\ref{equ:con_1}) holds only when the gradient of the outer objective with respect to $\alpha$ becomes 0 at some iteration t, \ie,
\begin{equation}
    {\mathcal{L}}^{weight}({\theta}_{t+1}) = {\mathcal{L}}^{weight}({\theta}_{t}),
\end{equation}
if and only if
\begin{equation}
    {\nabla}_{\alpha} \mathcal{L}^{weight}(\theta_t) = 0.
\end{equation}
\end{theorem}
\textbf{Proof.} Guo~\etal~\cite{guo2020safe} have proved the convergence of out-level objective. That is:
\begin{equation}
    \mathcal{L}^{outer}(\theta_{t+1}) - \mathcal{L}^{outer}(\theta_t) \leq 0,
\end{equation}
where $\mathcal{L}^{outer}$ is the out-level objective in bi-level optimization.

Therefore, the optimization objective of our weight model, which is the out-level objective, also converges during the iteration process:
\begin{equation}
    \mathcal{L}^{weight}(\theta_{t+1}) - \mathcal{L}^{weight}(\theta_t) \leq 0.
\end{equation}

\section{Datasets}
\label{sup:data}
We evaluate our method over nine public datasets, consisting of both natural image datasets and medical image datasets.
A detailed summary of the datasets we used is shown in Table~\ref{tab:dataset_statistics}.

The natural image datasets comprise Caltech 101, CIFAR100-Subset (CIFAR100-S), Standard Cars, Oxford 102 Flowers (Flowers), Oxford-IIIT Pets (Pets) and DTD. 
The CIFAR100-Subset dataset is derived from CIFAR100 by randomly selecting 100 images per class, yielding 10,000 instances across 100 classes.
This dataset serves as a representative example of a small-scale dataset for evaluating our method.
These natural datasets cover a variety of classification tasks, including coarse-grained object classification (\ie, Caltech 101 and CIFAR100-Subset), fine-grained object classification (\ie, Standard Cars, Flowers and Oxford-IIIT Pets), and textual classification (\ie, DTD).

The medical image datasets include BreastMNIST, PathMNIST and OrganSMNIST, which cover a wide range of medical image types, including breast ultrasound (\ie, BreastMNIST), colon pathology (\ie, PathMNIST), and abdominal CT (\ie, OrganSMNIST).
To simulate a small-scale dataset scenario, we used the validation sets of PathMNIST and BreastMNIST as training sets, while retaining the original training set for OrganSMNIST due to its smaller size.

\begin{table*}[t]
\centering
\caption{Statistics of both natural and medical image datasets.}
\label{tab:dataset_statistics}
\begin{tabular}{llccc}
\toprule
\textbf{Datasets} & \textbf{Type of  classification task} & \textbf{\# Classes} & \textbf{\# Samples} & \textbf{\# Average samples per class} \\ 
\midrule
Caltech 101 & Coarse-grained  & 102 & 3,060 & 30 \\
CIFAR100-Subset & Coarse-grained  & 100 & 10,000 & 100 \\
Standard Cars & Fine-grained  & 196 & 8,144 & 42 \\
Flowers & Fine-grained & 102 & 6,552 & 64 \\
Pets & Fine-grained  & 37 & 3,842 & 104 \\
DTD & Texture  & 47 & 3,760 & 80 \\ 
BreastMNIST & Medical (Breast Ultrasound) & 2 & 78 &39 \\
PathMNIST & Medical (Colon Pathology) & 9 & 10,004 &1,112 \\
OrganSMNIST & Medical (Abdominal CT) & 11 & 13,940 &1,267 \\
\bottomrule
\end{tabular}
\end{table*}

\section{More Results on Validating Correctness of the Theoretical Analysis}
\label{sup:ablation}
The experiments for validating the correctness of our theoretical analysis on the other three datasets are shown in Table~\ref{tab: ablation_organs}.
All these three datasets are configured with a $5 \times$ expansion ratio setting.
As we can see, all three datasets lead to similar conclusions in the manuscript: the strong connection supervision hurts the training process, but the pairwise connection supervision and the no connection supervision help.




\begin{table}[t]
    \centering
    \caption{Accuracy (in $\%$) of a ResNet-50 trained from scratch comparison of different combinations of the three types of supervision on DTD, Flowers and OrganSMNIST. }
        \begin{tabular}{ccccc}
            \toprule
             \multirow{2}{*}{dataset} & Strong & Pairwise & No & \multirow{2}{*}{Accuracy} \\
             &  connection & connection & connection & \\
            \midrule
            
             \multirow{7}{*}{DTD} & \checkmark &  &  & 37.2 \\ 
             \cmidrule(lr){2-5}
             & & \checkmark & &  41.1\\
             &\checkmark & \checkmark & & 33.2 ($\downarrow$ 7.9) \\ 
             \cmidrule(lr){2-5}
             & &  & \checkmark & 37.7 \\
             &\checkmark &  & \checkmark & 32.8 ($\downarrow$ 4.9) \\ 
             \cmidrule(lr){2-5}
             & & {\checkmark} & {\checkmark} & \textbf{50.3} \\
             &\checkmark & \checkmark & \checkmark & 34.6 ($\downarrow$ 15.7) \\

            \midrule
            \midrule
            
             \multirow{2}{*}{dataset} & Strong & Pairwise & No & \multirow{2}{*}{Accuracy} \\
             &  connection & connection & connection & \\
            \midrule
            
             \multirow{7}{*}{Flowers} & \checkmark &  &  & 82.1 \\ 
             \cmidrule(lr){2-5}
             & & \checkmark & & 82.5  \\
             &\checkmark & \checkmark & & 80.7 ($\downarrow$ 1.8) \\ 
             \cmidrule(lr){2-5}
             & &  & \checkmark & 83.2 \\
             &\checkmark &  & \checkmark & 81.8 ($\downarrow$ 1.4) \\ 
             \cmidrule(lr){2-5}
             & & {\checkmark} & {\checkmark} & \textbf{84.7} \\
             &\checkmark & \checkmark & \checkmark & 82.4 ($\downarrow$ 2.3) \\

            \midrule
            \midrule
            
             \multirow{2}{*}{dataset} & Strong & Pairwise & No & \multirow{2}{*}{Accuracy} \\
             &  connection & connection & connection & \\
            \midrule
            
               & \checkmark &  &  & 81.0 \\ 
             \cmidrule(lr){2-5}
             & & \checkmark & &  81.1 \\
             OrganS&\checkmark & \checkmark & & 80.1 ($\downarrow$ 1.0) \\ 
             \cmidrule(lr){2-5}
             MNIST& &  & \checkmark & 80.9 \\
             &\checkmark &  & \checkmark & 80.7 ($\downarrow$ 0.2) \\ 
             \cmidrule(lr){2-5}
             & & {\checkmark} & {\checkmark} & \textbf{81.2} \\
             &\checkmark & \checkmark & \checkmark & 80.3 ($\downarrow$ 0.9) \\

            \bottomrule

        \end{tabular}
    
    \label{tab: ablation_organs}
\end{table}


\section{More Results on the Correctness of the Image Accuracy}
\begin{figure*}[t]
  \centering
   \includegraphics[width=\textwidth]{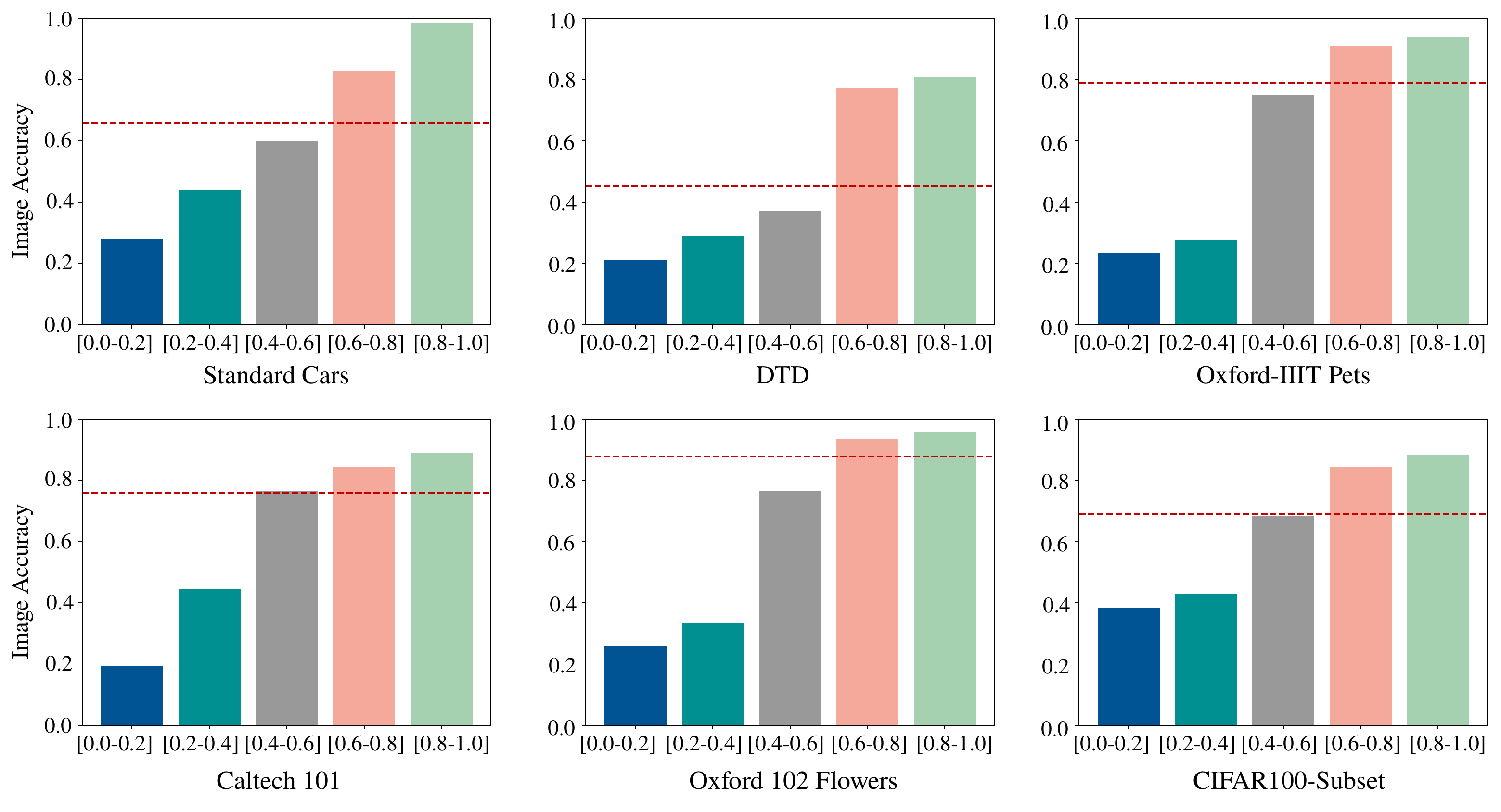}\Description{Graph showing image accuracy percentages for the dataset.}
   \caption{Image accuracy under different weights for all six natural datasets.}
   \label{fig:efficiency_nature}
\end{figure*}

We have demonstrated that our method effectively assigns low weights to noisy generated images on the Pets dataset in the manuscript.
Here, we further report the results on the other natural datasets.
As shown in Fig.~\ref{fig:efficiency_nature}, images with weights exceeding 0.6 consistently exhibit accuracy above the average across all six datasets.
These results reinforce that, regardless of the overall quality of the generated images, our method reliably assigns lower weights to noisy images and higher weights to clearer ones.

\begin{figure}[t]
  \centering
   \includegraphics[width=\linewidth]{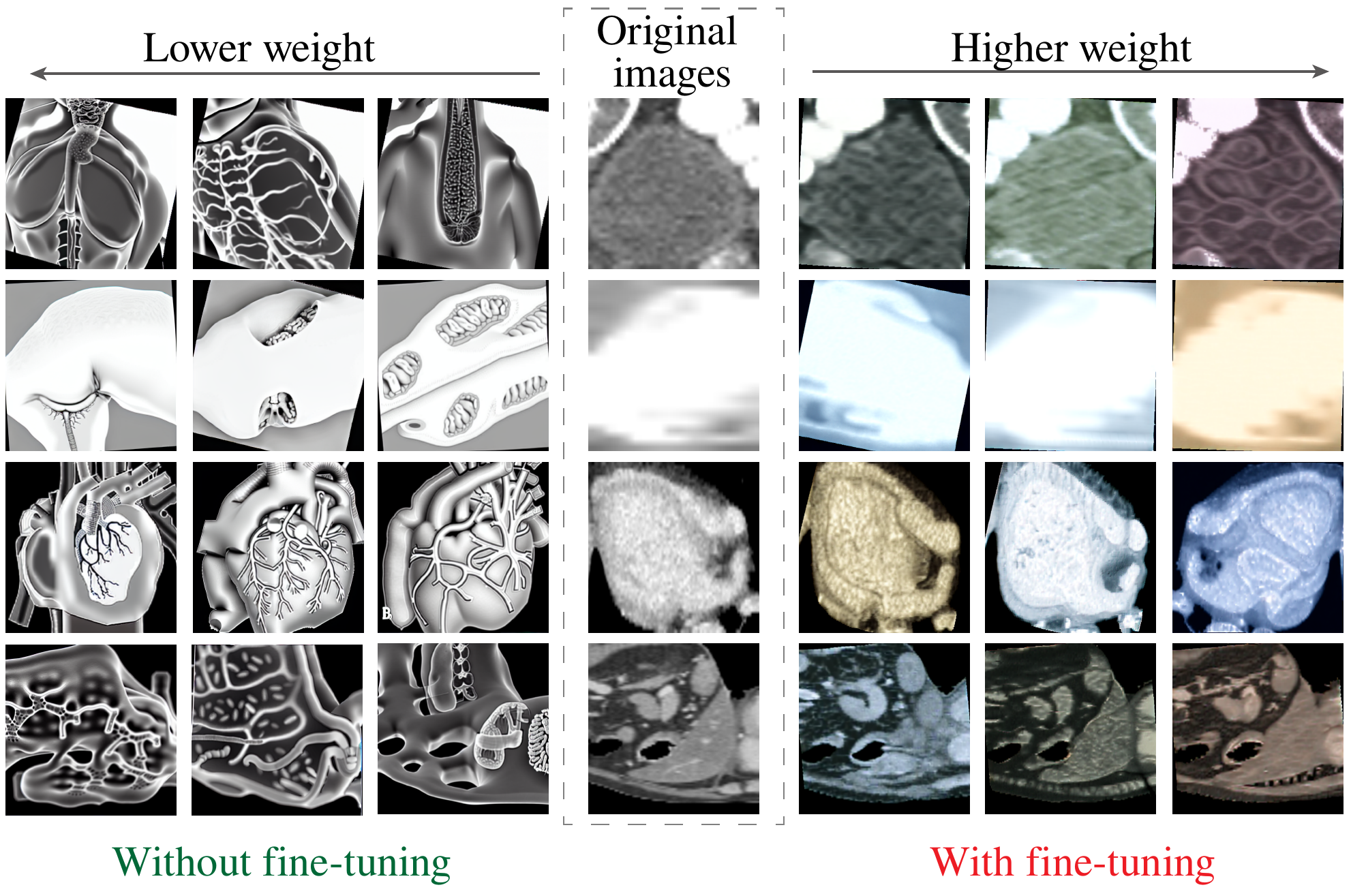}
   \caption{Exemplar original images on medical dataset and their associated generated images with and without fine-tuning ordered by their weights.}
   \label{fig:weight_medical}
\end{figure}

In Fig.~\ref{fig:weight_medical}, we also present medical images generated by GIF with and without fine-tuning, arranged according to their weights.
As we can see, without fine-tuning, GIF fails to produce images containing abdominal CT information for the OrganSMNIST dataset.
However, with fine-tuning, GIF generates similar yet distinct abdominal CT-related images, which enhance model performance.

\section{More Results on Generalization on Other SOTA Generative Models}
We have demonstrated the generalization ability of our method on the Pets dataset across different SOTA diffusion models in the manuscript.
Here, we present our experiments on the remaining eight datasets.
As shown in Table~\ref{Tab:all_sdmodels}, with $5 \times$-expanded datasets, ResNet-50 with our method achieves consistent improvements compared to without our method on all datasets.
This further demonstrates the generalization ability of TriReWeight.

Furthermore, for three medical datasets, training the model with a combination of expanded and original images led to a significant performance decline compared to training with only the original datasets.
This is because the images generated by diffusion models without fine-tuning are poor in quality, which hurts model performance. 
With TriReWeight, the accuracy is increased by a large gain, this further demonstrates that our method can effectively reduce the negative effect of noisy generated images.

\begin{table*}[t]
\centering
\caption{Model performance of training a ResNet-50 from scratch on $5 \times$-expanded datasets by different diffusion models with and without our method.}
\begin{tabular}{lcccc}
    \toprule
    \multirow{2}{*}{Dataset} & \multicolumn{4}{c}{\textbf{Caltech 101}}  \\
    \cmidrule(lr){2-5}
    & SD 1.4 (860M) & SD 2.1 (865M) &  SD 3.5 (8.1B) & SDXL (2.6B) \\
    \midrule
    w/o-TriReWeight & 45.33 & 55.68 & 56.76 & 59.81 \\
    w/-TriReWeight & \textbf{61.85}  (\textcolor{red}{+16.52}) & \textbf{65.69} 
 (\textcolor{red}{+10.01}) &  \textbf{68.70} (\textcolor{red}{+11.94}) & \textbf{71.3} (\textcolor{red}{+11.49}) \\
    \midrule
    
    \midrule
    \multirow{2}{*}{Dataset} & \multicolumn{4}{c}{\textbf{Cars}}  \\
    \cmidrule(lr){2-5}
    & SD 1.4 (860M) & SD 2.1 (865M) & SD 3.5 (8.1B) & SDXL (2.6B) \\
    \midrule
    w/o-TriReWeight & 41.87 & 64.45 & 62.33 & 66.40 \\
    w/-TriReWeight & \textbf{71.30}  (\textcolor{red}{+29.43}) & \textbf{75.84} 
 (\textcolor{red}{+11.39}) &  \textbf{79.85} (\textcolor{red}{+17.52}) & \textbf{85.39} (\textcolor{red}{+18.99}) \\
    \midrule

    \midrule
    \multirow{2}{*}{Dataset} & \multicolumn{4}{c}{\textbf{Flowers}}  \\
    \cmidrule(lr){2-5}
    & SD 1.4 (860M) & SD 2.1 (865M) &  SD 3.5 (8.1B) & SDXL (2.6B) \\

    \midrule
    w/o-TriReWeight & 78.48 & 80.68 & 81.54 & 77.75 \\
    w/-TriReWeight & \textbf{87.90}  (\textcolor{red}{+9.42}) & \textbf{89.24} 
 (\textcolor{red}{+8.56}) &  \textbf{91.93} (\textcolor{red}{+10.39}) & \textbf{90.35} (\textcolor{red}{+12.60}) \\
    \midrule

    \midrule
    \multirow{2}{*}{Dataset} & \multicolumn{4}{c}{\textbf{DTD}}  \\
    \cmidrule(lr){2-5}
    & SD 1.4 (860M) & SD 2.1 (865M) &  SD 3.5 (8.1B) & SDXL (2.6B) \\
    \midrule
    w/o-TriReWeight & 31.54 & 32.39 & 41.22 & 31.97 \\
    w/-TriReWeight & \textbf{47.07}  (\textcolor{red}{+15.53}) & \textbf{34.62} 
 (\textcolor{red}{+2.23}) &  \textbf{41.6} (\textcolor{red}{+0.38}) & \textbf{41.22} (\textcolor{red}{+9.25}) \\
    \midrule

    \midrule
    \multirow{2}{*}{Dataset} & \multicolumn{4}{c}{\textbf{CIFAR100-S}}  \\
    \cmidrule(lr){2-5}
    & SD 1.4 (860M) & SD 2.1 (865M) &  SD 3.5 (8.1B) & SDXL (2.6B) \\
    \midrule
    w/o-TriReWeight & 50.01 & 55.94 & 56.02 & 54.10 \\
    w/-TriReWeight & \textbf{55.32}  (\textcolor{red}{+5.31}) & \textbf{59.04} 
 (\textcolor{red}{+3.10}) &  \textbf{58.8} (\textcolor{red}{+2.78}) & \textbf{59.82} (\textcolor{red}{+5.72}) \\
    \midrule


    \midrule
    \multirow{2}{*}{Dataset} & \multicolumn{4}{c}{\textbf{PathMNIST}}  \\
    \cmidrule(lr){2-5}
    & SD 1.4 (860M) & SD 2.1 (865M) &  SD 3.5 (8.1B) & SDXL (2.6B) \\
    \midrule
    w/o-TriReWeight & 76.53 & 76.28 & 76.32 & 76.75\\
    w/-TriReWeight & \textbf{81.78}  (\textcolor{red}{+5.25}) & \textbf{85.12} 
 (\textcolor{red}{+8.84}) &  \textbf{84.97} (\textcolor{red}{+8.65}) & \textbf{84.81} (\textcolor{red}{+8.06}) \\
    \midrule

    \midrule
    \multirow{2}{*}{Dataset} & \multicolumn{4}{c}{\textbf{BreastMNIST}}  \\
    \cmidrule(lr){2-5}
    & SD 1.4 (860M) & SD 2.1 (865M) &  SD 3.5 (8.1B) & SDXL (2.6B) \\
    \midrule
    w/o-TriReWeight & 62.78 & 73.72 & 72.93 & 73.08 \\
    w/-TriReWeight & \textbf{69.23}  (\textcolor{red}{+6.45}) & \textbf{76.92} 
 (\textcolor{red}{+3.2}) &  \textbf{75.25} (\textcolor{red}{+2.32}) & \textbf{74.60} (\textcolor{red}{+1.52}) \\
    \midrule

    \midrule
    \multirow{2}{*}{Dataset} & \multicolumn{4}{c}{\textbf{OrganSMNIST}}  \\
    \cmidrule(lr){2-5}
    & SD 1.4 (860M) & SD 2.1 (865M) &  SD 3.5 (8.1B) & SDXL (2.6B) \\
    \midrule
    w/o-TriReWeight & 50.01 & 51.61 & 60.52 & 63.29 \\
    w/-TriReWeight & \textbf{78.93}  (\textcolor{red}{+28.92}) & \textbf{79.77} 
 (\textcolor{red}{+28.16}) &  \textbf{79.82} (\textcolor{red}{+19.30}) & \textbf{80.44} (\textcolor{red}{+17.15}) \\
    \bottomrule
\end{tabular}
\label{Tab:all_sdmodels}
\end{table*}

\begin{table*}[t]
\centering
\caption{Model performance of various backbones trained from scratch on $5 \times$-expanded datasets by GIF with and without our method.}
\begin{tabular}{lccccc}
    \toprule
    \multirow{2}{*}{\textbf{Dataset}} & \multicolumn{4}{c}{\textbf{Caltech 101}} \\
    \cmidrule(lr){2-6}
    & ResNet-50 & ResNeXt-50 & WideResNet-50 & MobileNet-v2 & \textbf{Average} \\
    \midrule
    wo-{\sys} & 54.4 & 52.8 & 60.7 & 55.6 & 55.9 \\
    w/-{\sys} & \textbf{76.7} & \textbf{69.3} & \textbf{70.8} & \textbf{70.5} & \textbf{71.8} (\textcolor{red}{+15.9}) \\
    \midrule
    
    \midrule
    \multirow{2}{*}{\textbf{Dataset}} & \multicolumn{4}{c}{\textbf{Cars}} \\
    \cmidrule(lr){2-6}
     & ResNet-50 & ResNeXt-50 & WideResNet-50 & MobileNet-v2 & \textbf{Average} \\
    \midrule
    wo-{\sys} & 60.6 & 64.1 & 75.1 & 60.2 & 65.0 \\
    w/-{\sys} & \textbf{73.4} & \textbf{77.9} & \textbf{79.4} & \textbf{78.5} & \textbf{77.3} (\textcolor{red}{+12.3}) \\
    \midrule

    \midrule
    \multirow{2}{*}{\textbf{Dataset}} & \multicolumn{4}{c}{\textbf{Flowers}} \\
    \cmidrule(lr){2-6}
     & ResNet-50 & ResNeXt-50 & WideResNet-50 & MobileNet-v2 & \textbf{Average} \\
    \midrule
    wo-{\sys} & 82.1 & 82.0 & 85.0 & 89.0 & 84.5 \\
    w/-{\sys} & \textbf{84.7} & \textbf{88.6} & \textbf{89.3} & \textbf{90.3} & \textbf{88.2} (\textcolor{red}{+3.7}) \\
    \midrule

    \midrule
    \multirow{2}{*}{\textbf{Dataset}} & \multicolumn{4}{c}{\textbf{DTD}} \\
    \cmidrule(lr){2-6}
     & ResNet-50 & ResNeXt-50 & WideResNet-50 & MobileNet-v2 & \textbf{Average} \\
    \midrule
    wo-{\sys} & 33.9 & 33.3 & 40.6 & 40.8 & 37.2 \\
    w/-{\sys} & \textbf{50.3} & \textbf{44.7} & \textbf{45.2} & \textbf{43.9} & \textbf{46.0} (\textcolor{red}{+8.8}) \\
    \midrule

    \midrule
    \multirow{2}{*}{\textbf{Dataset}} & \multicolumn{4}{c}{\textbf{CIFAR100-S}} \\
    \cmidrule(lr){2-6}
     & ResNet-50 & ResNeXt-50 & WideResNet-50 & MobileNet-v2 & \textbf{Average} \\
    \midrule
    wo-{\sys} & 61.1 & 59.0 & 64.4 & 62.4 & 61.7 \\
    w/-{\sys} & \textbf{62.8} & \textbf{67.6} & \textbf{68.9} & \textbf{69.7} & \textbf{67.3} (\textcolor{red}{+5.6}) \\
    \midrule


    \midrule
    \multirow{2}{*}{\textbf{Dataset}} & \multicolumn{4}{c}{\textbf{PathMNIST}} \\
    \cmidrule(lr){2-6}
     & ResNet-50 & ResNeXt-50 & WideResNet-50 & MobileNet-v2 & \textbf{Average} \\
    \midrule
    wo-{\sys} & 86.9 & 81.6 & 81.8 & 81.5 & 83.0 \\
    w/-{\sys} & \textbf{88.7} & \textbf{87.9} & \textbf{87.3} & \textbf{88.8} & \textbf{88.2} (\textcolor{red}{+5.2}) \\
    \midrule

    \midrule
    \multirow{2}{*}{\textbf{Dataset}} & \multicolumn{4}{c}{\textbf{BreastMNIST}} \\
    \cmidrule(lr){2-6}
     & ResNet-50 & ResNeXt-50 & WideResNet-50 & MobileNet-v2 & \textbf{Average} \\
    \midrule
    wo-{\sys} & 77.4 & 73.9 & 72.4 & 72.1 & 74.0 \\
    w/-{\sys} & \textbf{85.3} & \textbf{87.2} & \textbf{87.8} & \textbf{85.3} & \textbf{86.4} (\textcolor{red}{+12.4}) \\
    \midrule

    \midrule
    \multirow{2}{*}{\textbf{Dataset}} & \multicolumn{4}{c}{\textbf{OrganSMNIST}} \\
    \cmidrule(lr){2-6}
     & ResNet-50 & ResNeXt-50 & WideResNet-50 & MobileNet-v2 & \textbf{Average} \\
    \midrule
    wo-{\sys} & 80.7 & 70.9 & 73.7 & 75.2 & 75.1 \\
    w/-{\sys} & \textbf{81.2} & \textbf{82.2} & \textbf{81.4} & \textbf{81.9} & \textbf{81.7} (\textcolor{red}{+6.6}) \\
    \bottomrule
\end{tabular}
\label{Tab:all_backbones}
\end{table*}

\section{More Results on Generalization to Other Classification Model Backbones}

We have demonstrated the generalization ability of our method across various backbones using the Pets dataset in the manuscript.
Here, we present the experiments on the remaining eight datasets.
These datasets were generated by performing $5 \times$ expansion ratios of GIF and combined with the original ones to train ResNeXt-50, WideResNet-50, and MobileNet-v2 from scratch, with and without our method.
As shown in Table~\ref{Tab:all_backbones}, our method surpasses GIF across all backbones for all datasets. 
These findings demonstrate the robust generalization ability of {\sys}, which can be applied to diverse backbones within the same training framework.

\end{document}